\documentclass[a4paper,fleqn]{cas-dc}

\usepackage[numbers,sort&compress]{natbib}
\usepackage{amsmath}
\usepackage{amssymb}
\usepackage{array}
\usepackage{graphicx}
\usepackage{xurl}
\usepackage{placeins}
\usepackage[ruled,linesnumbered,vlined]{algorithm2e}

\AtBeginDocument{\setlength{\mathindent}{0pt}}

\makeatletter
\g@addto@macro\normalsize{%
  \setlength{\abovedisplayskip}{5pt plus 1pt minus 1pt}%
  \setlength{\belowdisplayskip}{5pt plus 1pt minus 1pt}%
  \setlength{\abovedisplayshortskip}{3pt plus 1pt minus 1pt}%
  \setlength{\belowdisplayshortskip}{4pt plus 1pt minus 1pt}%
  \setlength{\jot}{2pt}%
}
\makeatother


\newcolumntype{P}[1]{>{\raggedright\arraybackslash}p{#1}}
\newcommand{\compacttablesetup}{%
  \fontsize{6.4pt}{7.1pt}\selectfont
  \renewcommand{\arraystretch}{0.98}%
  \setlength{\tabcolsep}{2pt}%
}

\newcommand{\toolname}{HCRMap}

\newcommand{\dtd}{D2D}

\newcommand{\edp}{EDP}
\newcommand{\threed}{3.5D}

\RenewDocumentCommand{\printorcid}{}{}
\ExplSyntaxOn
\bool_gset_true:N \g_stm_nologo_bool
\ExplSyntaxOff

\begin{document}
\let\WriteBookmarks\relax

\shorttitle{Pressure-Aware Residency Mapping for 3.5D MoE}
\shortauthors{Y. Zhang}

\title[mode=title]{\toolname{}: Pressure-Aware Hot-Expert Residency Mapping\\ for \threed{} MoE Chiplet Inference}

\author[1]{Yongqin Zhang}
\cormark[1]
\ead{zyq1350335@126.com}

\affiliation[1]{
  organization={Nanjing Vocational College of Information Technology},
  addressline={99 Wenlan Road, Xianlin College Community},
  city={Nanjing},
  state={Jiangsu},
  postcode={210023},
  country={P.R.C.}
}

\cortext[cor1]{Corresponding author.}

\begin{abstract}
Mixture-of-Experts (MoE) large language models (LLM) activate only a small number of experts during inference, but token routing introduces persistent expert hotness skew: a small set of hot experts continuously receives most tokens, while the remaining experts are lightly loaded. On 3.5D multi-chiplet systems, this skew not only causes compute imbalance but also amplifies pressure on communication, memory bandwidth, I/O, and execution queues. Therefore, the core problem is not simply to reduce token movement, but to dynamically place and reuse hot expert replicas across different memory tiers.

This paper proposes HCRMap, a hot expert residency mapping framework for pressure-aware expert replica management in 3.5D MoE inference. Based on expert hotness, weight loading cost, migration overhead, and runtime resource pressure, HCRMap dynamically determines which experts should be promoted, retained, demoted, or evicted. It then maps routed token groups to suitable resident replicas, thereby jointly mitigating communication, memory, and queue bottlenecks.

Experimental results show that HCRMap reduces end-to-end latency by 43.6\% and 43.0\% over Hydra in the prefill and decode stages, respectively; by 34.5\% and 33.1\% over MoEntwine; and by 46.7\% and 46.0\% over PIMoE.
\end{abstract}



\maketitle

\section{Introduction}
\label{sec:introduction}

\begin{figure}[pos=t,align=\centering]
\centering
\includegraphics[width=\columnwidth]{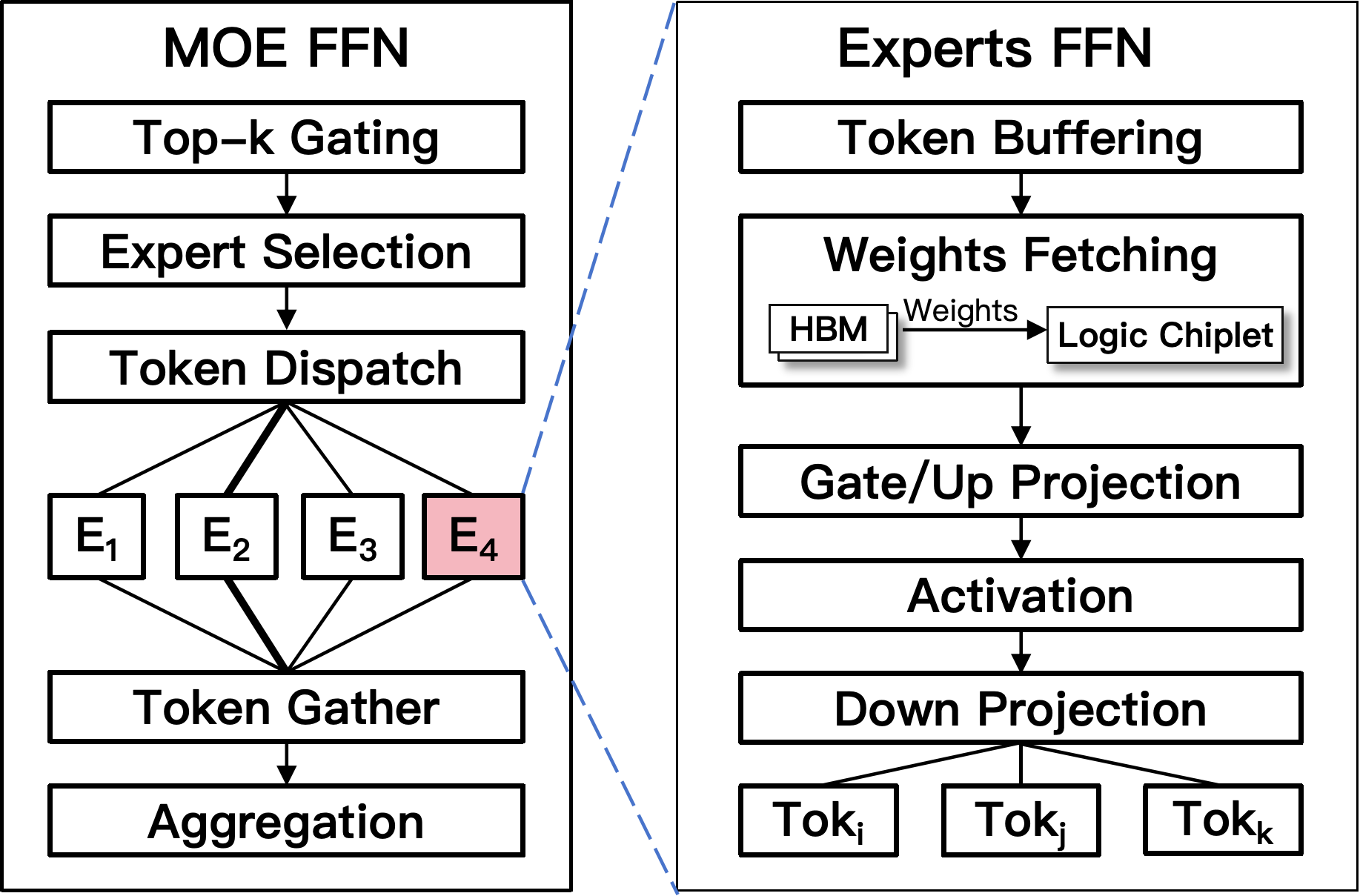}
\caption{MoE FFN serving pipeline for selected experts.}
\label{fig:moe-arch-pipeline}
\end{figure}

Mixture-of-Experts large language models (MoE-LLMs) scale capacity by activating only a few experts per token through Top-$k$ gating~\cite{fedus2022switch,jiang2024mixtral,deepseekmoe2024,qwen32025}, but token-dependent routing produces persistent expert-popularity skew: a small set of hot experts repeatedly receives most tokens across serving windows while others stay lightly used~\cite{hwang2023tutel,he2026hydra,zhu2026probe,wei2026ultraep}. As shown in Figure~\ref{fig:moe-arch-pipeline}, this skew is not only a compute-load problem. It reshapes token dispatch and gather, expert-weight access, memory-bank pressure, and shared-I/O pressure. The routed expert-FFN bottleneck is therefore a cross-resource imbalance over compute, memory, and interconnect, affecting full-model latency and energy in both prefill and decode execution.

On \threed{} multi-chiplet systems with hierarchical memory, this skew surfaces as repeated expert-weight streaming~\cite{wang2025lamosys,luo2025mozart,ma2026expertstreaming}. Expert weights are too large to keep all active experts in the nearest stacked-SRAM tier; in our evaluation, each benchmark uses the routed-expert footprint computed from its own hidden and intermediate dimensions, while the near-tier budget of each chiplet group is only 64 MB. Therefore, finite near-tier capacity can hold only a small number of hot experts, while routing profiles may require more hot or warm resident experts in the same layer. Warm experts that overflow the near tier must repeatedly stream from group-shared DRAM through shared I/O and D2D/NoP links. This transfer recurs across windows and continuously occupies shared interfaces and package links. Because \threed{} integration co-packages stacked SRAM, local HBM, and group-shared DRAM into a finer-grained hierarchy, expert weights should not be classified simply as resident or non-resident; they can be placed at different residency levels according to runtime hotness, streaming cost, and resource pressure.

Existing MoE systems optimize expert execution along separate axes: placement and routing that use popularity, affinity, or co-activation to cut dispatch latency; replication or shadow replicas that relieve hot-expert queueing; and heterogeneous execution or memory-aware offloading driven by compute and memory behavior~\cite{zhu2026probe,wei2026ultraep,zhao2026craft,han2025gracemoe,yu2026finemoe,li2025diffmoe,kim2024monde}. Hydra~\cite{he2026hydra}, MoEntwine~\cite{tang2026moentwine}, and PIMoE~\cite{wu2026pimoe} respectively target popularity-aware remapping, wafer-scale shadow replication, and NPU--PIM execution, while Mozart~\cite{luo2025mozart} shows the potential of profile-driven allocation and hierarchical streaming on a \threed{} base. Several of these systems combine multiple signals: PROBE co-balances computation and communication with dynamic replication, GRACE-MoE couples replication with locality-aware routing, and MoEntwine hides migration over cold links. Their shared limitation is that these signals are not tied to a \threed{} MoE layer-latency model that includes tier capacity, memory-bank service, shared-I/O pressure, and live package-link occupancy. That gap leaves open the question that governs \threed{} MoE performance: where hot expert weights reside, which copy a token is sent to, and how those choices propagate into dispatch, queueing, memory, compute, and gather latency.

Our characterization with LEGOSim~\cite{lin2025legosim} makes this concrete. First, minimizing communication does not minimize latency: the reproduced Hydra baseline substantially reduces D2D traffic but leaves most end-to-end latency unresolved, because reduced token movement does not relieve compute imbalance, repeated streaming, or bank pressure. Second, resident-copy capacity creates a sharp boundary: when the hot set cannot receive enough execution locations, latency increases sharply relative to the capacity-sufficient setting. Separately, an intermediate HBM tier reduces repeated streaming when warm experts do not fit in the nearest tier. Hot-expert residency is thus not a cache-hit problem but a cross-resource tradeoff: it spends near-tier capacity, copy bandwidth, and token-routing flexibility to reduce repeated streaming, queueing, and memory pressure. Traffic is both an overhead and a means of moving work to a better tier, so it should be budgeted and scheduled as part of latency control. Concretely, ``scheduling traffic'' means four runtime decisions that Section~\ref{sec:method} develops: which resident copy a routed token is sent to, whether an expert weight is promoted or demoted across tiers, whether a cross-tier migration is paced or deferred, and whether that migration avoids a dispatch- or gather-congested link.

Realizing this control on a \threed{} package is constrained on several fronts. Residency tiers differ in capacity, path, and sharing scope: SRAM is cheapest but tightest and must reserve space for activations and buffers; local HBM adds capacity but finite bank bandwidth; shared DRAM has the most capacity but repeatedly consumes shared I/O and NoP paths. Promotion and demotion are not free; they contend with foreground dispatch and gather. A token's completion time depends on target queue, bank, path, and shared-I/O pressure, so the shortest distance or lowest load need not yield the lowest completion time. Finally, hotness drifts with serving mode, batch composition, and request domain: reacting to every burst churns migrations before copy cost amortizes, while reacting too slowly leaves new hot experts remote. Stable online control therefore requires minimum residency, benefit-cost amortization, hysteresis, and bounded updates.

We propose \toolname{}, a two-timescale pressure-aware residency-mapping framework on a Mozart \threed{} base~\cite{luo2025mozart}. It adds a bounded hot-expert residency substrate--bank-isolated replica regions, a versioned cross-tier expert-to-replica directory, non-blocking chunked weight transfer, and lightweight pressure counters for compute, links, banks, and shared I/O--so that warm experts can be demoted to an intermediate tier instead of being evicted to far DRAM. A masked feasibility-constrained slow residency loop decides promotion, retention, demotion, and eviction across tiers, weighing streaming and queueing reduction against copy overhead, token-transfer cost, destination pressure, tier capacity, and residency age; a deterministic fast token loop assigns tokens to the current layout under queueing, path, bank, and streaming-cost pressure. Separating the two timescales keeps the layout stable while routing reacts to instantaneous pressure. The slow controller uses fixed checkpoints and legal action masks, while the fast loop uses fixed pressure-proxy weights.

Under matched resident-copy capacity, the latency-oriented placement and token assignment of \toolname{} reduce normalized end-to-end latency by 24.95\% and 23.91\% relative to the strongest replication-capable baseline in prefill and decode, respectively, and improve EDP by 19.94\% and 19.86\%. In the unified LEGOSim evaluation, where Hydra, MoEntwine, PIMoE, and \toolname{} run on the same \threed{} substrate, \toolname{} reduces normalized end-to-end latency by 34.5\% and 33.1\% relative to the strongest reproduced baseline in prefill and decode, respectively. This paper makes two contributions:

\begin{itemize}
  \item \textbf{A bounded multi-level hot-expert residency substrate for \threed{} MoE serving.} The substrate distributes hot-expert weights across stacked SRAM, chiplet-local HBM, and group-shared DRAM by streaming cost and hotness, letting warm experts degrade to an intermediate tier and avoid the near-tier capacity cliff.
  \item \textbf{A two-timescale pressure-aware mapper.} The mapper couples slow cross-tier residency updates with fast token assignment, using minimum residency, amortization, hysteresis, and bounded updates for stable online control.
\end{itemize}

The rest of this paper is organized as follows. Section~\ref{sec:related} reviews related work; Section~\ref{sec:motivation} motivates the pressure-aware residency problem; Section~\ref{sec:system} defines the system model; Section~\ref{sec:formulation} formulates the latency-minimization mapping problem; Section~\ref{sec:method} presents \toolname{}; Section~\ref{sec:experiments} evaluates the design; and Section~\ref{sec:conclusion} concludes.

\section{Related Work}
\label{sec:related}

This section reviews the work most related to \toolname{}. These systems already use expert popularity, replication, prefetching, cold-expert offloading, or dynamic balancing to handle hot experts. Their key difference is which latency bottleneck they expose to the runtime mapper when deciding whether a residency update, expert placement, or token reassignment is worth performing.

\subsection{Runtime Expert Balancing and Replication}

Tutel addresses distributed training and serving challenges caused by routing and load imbalance~\cite{hwang2023tutel}. Production serving stacks also expose similar mechanisms; for example, TensorRT-LLM documents Wide-EP and EPLB deployment choices for large MoE models~\cite{nvidia2026wideep}.

More recent work makes replication and balancing more explicit. PROBE co-balances computation and communication through lookahead prediction, dynamic replication, and token assignment~\cite{zhu2026probe}. UltraEP performs exact-load, real-time balancing for expert-parallel training and prefill serving on rack-scale nodes~\cite{wei2026ultraep}. CRAFT studies fine-grained expert replication under a memory budget and shows that naive replication can over-replicate experts with little benefit~\cite{zhao2026craft}. GRACE-MoE combines grouping, dynamic replication, and locality-aware routing for distributed MoE inference~\cite{han2025gracemoe}.

These systems all treat hot experts as dynamic runtime objects. Token splitting and migration hiding are also not new by themselves: GRACE-MoE uses locality-aware replica selection, and MoEntwine's NI-Balancer performs non-invasive migration through cold links~\cite{han2025gracemoe,tang2026moentwine}. They differ from \toolname{} in the hardware cost model. In GPU clusters, rack-scale nodes, or WSC meshes, the main costs are rank load, all-to-all communication, GPU memory, topology-specific cold links, and exposed transfer overhead. In \threed{} chiplet inference, the same action must also account for package-link pressure, switch/link occupancy that overlaps with dispatch and gather, shared HBM-bank or I/O pressure, and near-tier residency capacity. These costs do not appear in GPU or rack-scale cost models.

\subsection{Hot/Cold Expert Tiering, Caching, and Streaming}

FineMoE uses fine-grained expert selection patterns and semantic hints to guide expert prefetching, caching, and offloading for MoE serving~\cite{yu2026finemoe}. DiffMoE uses priority-driven differential expert caching for batched MoE inference~\cite{li2025diffmoe}. MoNDE transfers hot experts to the GPU and computes cold experts near host memory~\cite{kim2024monde}. Expert Streaming proposes FSE-DP for low-batch MoE inference on multi-chiplet accelerators, orchestrating fine-grained expert streams across D2D links~\cite{ma2026expertstreaming}. These systems mainly expose cache misses, offload delay, or streaming delay to the scheduler.

These works do not make the same mapping decision as \toolname{}. A cache/offload system asks which expert weights should reside in which memory tier. Expert Streaming asks how complementary expert streams should move along a trajectory. These decisions do not jointly decide expert residency and token routing across multiple residency levels under real-time physical pressure. Cache hit rate is therefore also insufficient for this problem: it only says whether a weight is near the consumer, whereas \threed{} residency control is multi-level, multi-replica, and coupled to token routing.

\subsection{Chiplet and Wafer-Scale MoE Accelerators}

Another line of work organizes MoE execution directly on chiplet or wafer-scale substrates. These systems are the closest targets used in our evaluation.

Hydra targets chiplet MoE accelerators~\cite{he2026hydra}. It predicts expert popularity and maps popular experts close to dominant token sources to reduce routed-token D2D communication. Popularity and source distance can provide a good initial layout, but they do not account for how a multi-level residency update changes queueing, bank-service, and migration latency under live link pressure, bank pressure, and tier-capacity pressure. A distance-optimal location may still contend with dispatch and gather on the same D2D link.

MoEntwine targets wafer-scale chips~\cite{tang2026moentwine}. It uses ER-Mapping to balance WSC communication and NI-Balancer to perform non-invasive expert migration over cold links. Its cost model depends on wafer-mesh topology and the availability of complementary cold links. This abstraction emphasizes WSC topology, while \threed{} chiplet inference also exposes tier and bank pressure.

PIMoE targets NPU-PIM systems~\cite{wu2026pimoe}. It uses throttle-aware task offloading to send memory-bound cold-expert work toward PIM while keeping hotter work on NPU resources. It therefore centers execution-medium selection, while \threed{} residency control couples dispatch, queueing, memory service, and gather latency across residency levels. For evaluation, we implement the expert-management policies of Hydra, MoEntwine, and PIMoE inside the same LEGOSim event flow used by \toolname{}. This setup evaluates all policies under identical \threed{} hardware constraints, routing outputs, memory hierarchy, and energy accounting.

Mozart is the closest \threed{} MoE architecture reference~\cite{luo2025mozart}. It provides chiplet modularity, co-activated expert allocation, token and expert streaming, and a NoP-tree topology for MoE training. However, Mozart generates expert allocation from offline profiles, targets training-oriented static layout, and does not manage multi-level residency during inference under real-time physical pressure. LaMoSys3.5D studies a scalable 3.5D-IC architecture for LLM serving, co-designing heterogeneous 3D-DRAM chiplets, dataflow, and scheduling~\cite{wang2025lamosys}. These works show that \threed{} platforms are plausible for large-model serving, but they do not manage inference-time multi-level residency updates as runtime decisions.

Recent Tau ($\tau$) Scaling Law work, also known as Tao's Law, further frames post-Moore scaling as time-constant reduction across devices, circuits, chips, and systems~\cite{he2026timescaling}. \toolname{} is orthogonal to this device/circuit-level direction: it does not introduce new 3D devices or bonding structures, but studies how a \threed{} MoE runtime can reduce the effective system-level time spent on expert-weight movement, bank service, package-link contention, and hot-expert queues.

\section{Motivation: Challenges and Opportunities in \threed{} MoE Serving}
\label{sec:motivation}

When MoE inference runs on a \threed{} (2.5D/3D hybrid-packaged) multi-chiplet system, the problem caused by hot experts is no longer only conventional compute-load imbalance. Because MoE uses token-dependent routing, a small number of hot experts repeatedly receive many tokens across consecutive serving windows. In ordinary distributed systems, this phenomenon is usually understood as expert-load imbalance or communication imbalance. In a \threed{} package, however, the same hot routing pattern simultaneously determines which \dtd{} or NoP path each token takes, which memory tier supplies the expert weights, which memory banks are accessed, and whether shared I/O remains continuously occupied. Therefore, the hot-expert problem in a \threed{} system evolves into a coupled pressure problem across compute, memory, and interconnect.

\begin{figure*}[width=\textwidth,pos=t,align=\centering]
\centering
\includegraphics[width=\textwidth]{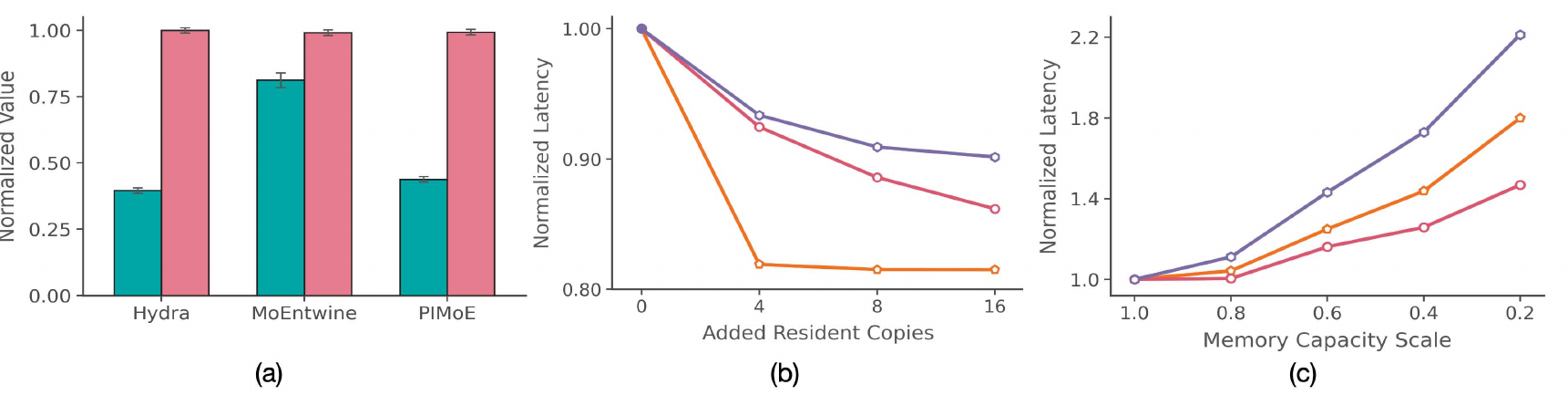}
\caption{Motivation observations from LEGOSim on the shared \threed{} substrate: (a) communication-oriented placement; (b) added resident copies; (c) reduced memory capacity.}
\label{fig:motivation-residency-sweep}
\end{figure*}

\begin{figure}[pos=t,align=\centering]
\centering
\includegraphics[width=\columnwidth]{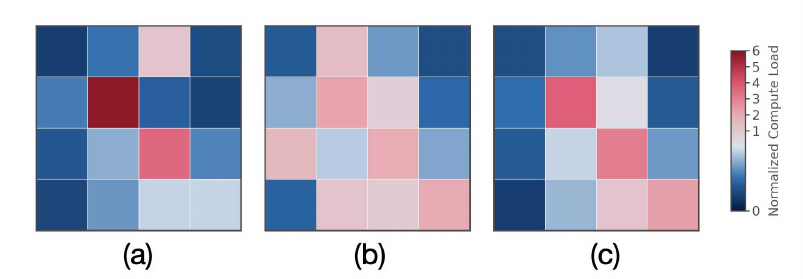}
\caption{Mechanism sketches: (a) residual hot-load concentration; (b) replica-enabled load spreading; (c) capacity-constrained re-concentration.}
\label{fig:motivation-three-panel}
\end{figure}

\textbf{Challenge 1.} \threed{} provides a richer hierarchical memory structure, but the resource attributes of different tiers are not the same. SRAM is closest to the compute units and has the lowest access latency, so it is suitable for hot experts, but its capacity is limited and it must also reserve space for activations, buffers, and metadata. HBM provides larger intermediate capacity and is suitable for warm experts or demoted hot experts, but its bank bandwidth is limited and can easily form a memory-service bottleneck. DRAM provides the largest shared capacity, but repeatedly streaming expert weights from DRAM continuously occupies shared I/O and \dtd{}/NoP paths. Therefore, placing a hot expert at the location closest to the token source does not necessarily minimize end-to-end latency; that location may still be disturbed by compute queueing, bank contention, or dispatch/gather traffic.

Figure~\ref{fig:motivation-residency-sweep}(a) shows this challenge. Taking directly simulated communication-oriented one-copy baselines as an example, these methods can significantly reduce \dtd{} traffic, but the final latency decreases only slightly. They mainly shorten the token transfer path while leaving the hot expert's service location unchanged. As shown in Figure~\ref{fig:motivation-three-panel}(a), tokens still queue behind a small number of overloaded expert locations, and the system critical path remains dominated by residual hot-load concentration instead of average communication distance.

\textbf{Challenge 2.} In a \threed{} system, expert replication is not unconditionally effective. Figure~\ref{fig:motivation-residency-sweep}(b) shows that adding a small number of resident copies can significantly reduce latency, because each resident copy is not only redundant weight storage, but also an additional legal execution slot for a hot expert. It can spread tokens that were originally concentrated in one long queue across multiple near execution locations, thereby alleviating hot-expert queueing.

However, Figure~\ref{fig:motivation-residency-sweep}(c) further shows that when memory capacity decreases, the benefit of fixed replication quickly disappears and may even increase latency. As shown in Figure~\ref{fig:motivation-three-panel}(c), when near-tier capacity is insufficient, some expert copies cannot continue to reside in useful locations, or they must compete for service bandwidth from the same memory bank. The load that was originally spread by replication re-concentrates on a small number of feasible locations, while replication itself introduces additional capacity occupation, weight movement, and memory-service pressure. Therefore, a resident copy is not simply one more stored weight copy, but a scarce near-execution resource; it truly helps reduce latency only when the system can retain and serve it at low cost.

These phenomena together reveal the core problem in \threed{} MoE serving: hot-expert management is not a pure communication-minimization problem, not a simple replication-maximization problem, and not something that a conventional cache hit rate can fully describe. The real problem is pressure-aware residency scheduling. The runtime system must simultaneously decide which hot experts deserve extra resident locations, which memory tier and memory bank should hold these copies, which expert copy each routed token should be sent to, and whether promotion, demotion, or migration should be suppressed when capacity, bank, link, or shared-I/O pressure becomes too high.

At the same time, \threed{} also brings a new opportunity. Unlike a flat accelerator memory model, \threed{} naturally provides a multi-level residency hierarchy composed of SRAM, HBM, and DRAM. This means expert weights do not need to be simply divided into resident and non-resident states. The hottest experts can occupy scarce SRAM regions; warm experts can be retained in or demoted to HBM to avoid repeated streaming from shared DRAM; and cold experts can stay in the shared capacity tier. This multi-level residency path provides new scheduling freedom for the runtime system, allowing it to trade near-tier capacity, copy bandwidth, and token-routing flexibility in order to reduce repeated weight movement, hot-expert queueing, and memory pressure.

Based on this motivation, this paper proposes \toolname{}. \toolname{} does not treat expert copies as redundancy that should be increased unconditionally, but as constrained residency resources. Through a slow residency loop, it performs promotion, retention, demotion, and eviction across memory tiers while considering hotness stability, saved streaming cost, queueing reduction, migration exposure, tier capacity, and bank pressure. Through a fast token-mapping loop, it selects an appropriate resident copy for each routed token group under the current residency layout, balancing queue pressure, path pressure, bank pressure, and streaming cost. By separating stable cross-tier residency updates from instantaneous token-assignment decisions, \toolname{} uses the multi-level resource opportunity provided by \threed{} systems while avoiding replication and migration themselves becoming new bottlenecks.

\section{System Model}
\label{sec:system}

\subsection{Software Execution Model}

\textbf{Software execution graph.} LEGOSim instantiates inference events for each evaluated MoE model, including non-routed transformer events and routed MoE-FFN events. The formal model captures the routed-MoE matching problem among routed token groups, expert-weight objects, resident expert replicas, and token-to-replica assignments. Let $\mathcal{L}$ be the set of modeled MoE layers. Layer $l\in\mathcal{L}$ has an expert set $\mathcal{E}_l$, and its routed token groups are
$\mathcal{G}_l=\{g\mid g=(l,e_g,s_g,a_g,N_g)\}$
Here $e_g\in\mathcal{E}_l$ is the expert selected by the router for group $g$, $s_g\in\mathcal{C}$ is the source compute chiplet that holds the input activations, $a_g\in\mathcal{C}$ is the aggregation chiplet for the expert output, and $N_g$ is the number of tokens in the group. The compute-chiplet set $\mathcal{C}$ is defined by the hardware model below.

Expert $e$ in layer $l$ has a weight object $W_{l,e}$ with size $S_{l,e}^{\mathrm{wt}}$. At runtime, the expert can have one or more resident replicas $\mathcal{R}_{l,e}$, and all replicas in the layer are
$\mathcal{R}_l=\bigcup_{e\in\mathcal{E}_l}\mathcal{R}_{l,e}$
For a replica $r$, $\ell(r)$ denotes its layer and $e(r)$ denotes its logical expert. The token-assignment variable is $x_{g,r}\in\{0,1\}$, where $x_{g,r}=1$ means token group $g$ executes on resident replica $r$. Routing correctness requires each group to choose exactly one legal replica of its target expert:
$\sum_{r\in\mathcal{R}_{l,e_g}} x_{g,r}=1,\ \forall l\in\mathcal{L},\ g\in\mathcal{G}_l$

\subsection{Hardware Resource Model}

\begin{figure}[pos=t,align=\centering]
\centering
\includegraphics[width=\columnwidth]{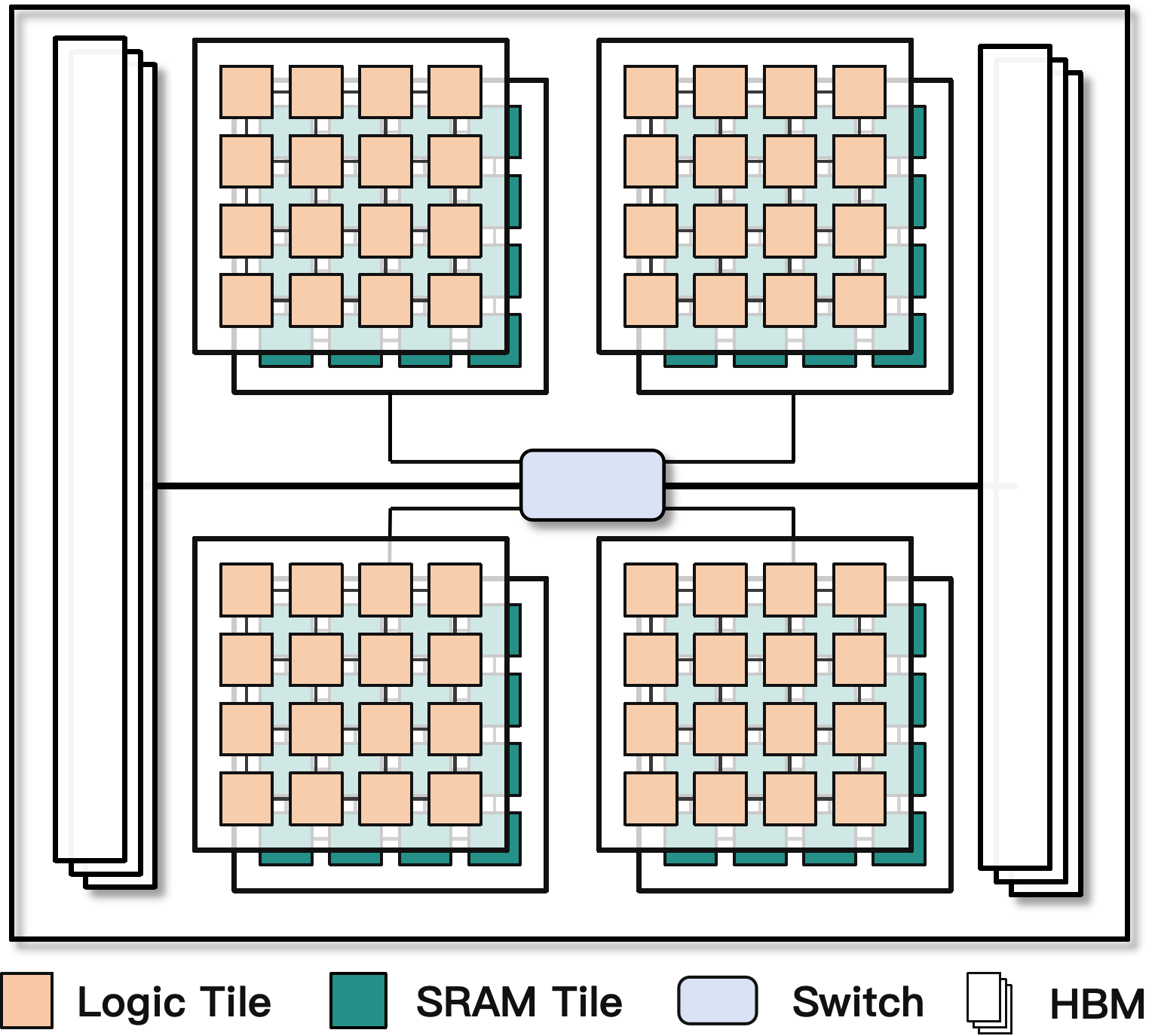}
\caption{Target \threed{} multi-chiplet substrate.}
\label{fig:hardware-substrate}
\end{figure}

Figure~\ref{fig:hardware-substrate} illustrates the \threed{} multi-chiplet substrate used by the model. The figure grounds the abstractions used below: logic tiles form the compute-chiplet set $\mathcal{C}$, local SRAM/HBM/DRAM regions form tiered bank sets $\mathcal{B}_{t,c}$, and mesh links, switch-fabric paths, \dtd{} links, and memory-access paths form the hardware edge set $E_{\mathrm{hw}}$.

The accelerator is represented as a directed hardware graph
$G_{\mathrm{hw}}=(V_{\mathrm{hw}},E_{\mathrm{hw}})$
The node set contains compute chiplets, routers, memory controllers, and memory banks. Each link $u\in E_{\mathrm{hw}}$ has bandwidth $B_u$, per-byte energy $\epsilon_u$, physical distance $\delta_u$, signal speed $v_u$, and fixed routing overhead $\lambda_u^0$. Its fixed transfer latency is
$\lambda_u=\lambda_u^0+\delta_u/v_u$
Each compute chiplet $c\in\mathcal{C}$ has $N_c^{\mathrm{core}}$ cores, per-core throughput $\Pi_{\mathrm{core}}$, aggregate throughput $\Pi_c$, and buffer capacity $S_c^{\mathrm{buf}}$:
$\Pi_c=N_c^{\mathrm{core}}\Pi_{\mathrm{core}}$

Expert weights can reside in three tiers
$\mathcal{T}=\{\mathrm{SRAM},\mathrm{HBM},\mathrm{DRAM}\}$
SRAM provides near hot-expert residency, HBM provides intermediate capacity for warm or demoted hot experts, and DRAM provides shared capacity and fallback residency. For tier $t\in\mathcal{T}$ and chiplet $c\in\mathcal{C}$, $\mathcal{B}_{t,c}$ denotes the visible bank set. Bank $(t,c,b)$ has capacity $Cap_{t,c,b}$, bandwidth $BW_{t,c,b}$, fixed access latency $Lat_{t,c,b}$, and per-byte access energy $\epsilon_{t,c,b}$. All evaluated policies use the same graph, tier capacities, bank bandwidths, link bandwidths, and compute resources.

\subsection{Expert Residency and Constraints}

Replica $r$ is placed by
$\phi(r)=(t_r,c_r,b_r)$
where $t_r\in\mathcal{T}$ is its tier, $c_r\in\mathcal{C}$ is its execution chiplet, and $b_r\in\mathcal{B}_{t_r,c_r}$ is its backing bank. For any physical bank, resident weights, activation/token-buffer reservation, and metadata state must fit:
\begin{equation}
\sum_{r:\phi(r)=(t,c,b)}
S_{\ell(r),e(r)}^{\mathrm{wt}}
+ A_{t,c,b}+M_{t,c,b}
\le Cap_{t,c,b}.
\end{equation}
Here $A_{t,c,b}$ reserves space for activations and token buffers, and $M_{t,c,b}$ accounts for the directory, version state, and replication buffers. Replica counts and the global residency budget satisfy
\begin{equation}
\begin{aligned}
1 \le |\mathcal{R}_{l,e}| \le R_{l,e}^{\max}, \qquad
\sum_{l\in\mathcal{L}}\sum_{e\in\mathcal{E}_l}
S_{l,e}^{\mathrm{wt}}|\mathcal{R}_{l,e}|
\le B_{\mathrm{res}} .
\end{aligned}
\end{equation}

The runtime directory keeps the logical-to-physical mapping
$\mathcal{D}(l,e)=\{(r,\phi(r),\nu_r)\mid r\in\mathcal{R}_{l,e}\}$
where $\nu_r$ is the published version. During promotion or demotion, the target chunks are copied into a private pending region. The mapper publishes a new directory version only after the copy completes; token groups already assigned to the old version continue there until their in-flight count reaches zero. Because inference weights are read-only, this versioned protocol prevents partial-copy reads without requiring coherence between old and new replicas.

Let $\xi_r\in\{0,1\}$ indicate whether replica $r$ is used in the current layer execution. Token assignment and replica use are coupled by
\begin{equation}
x_{g,r}\le \xi_r,\qquad
\xi_r\le
\sum_{g\in\mathcal{G}_l:e_g=e(r)}x_{g,r}.
\end{equation}
Let $k_{r,c}\in\mathbb{Z}_{\ge0}$ be the number of cores assigned to replica $r$ on chiplet $c$. Since a replica executes on its mapped chiplet, $k_{r,c}=0$ for $c\ne c_r$, and
\begin{equation}
\begin{aligned}
\xi_r \le k_{r,c_r} \le N_{c_r}^{\mathrm{core}}\xi_r,\qquad
\sum_{r\in\mathcal{R}_l:c_r=c} k_{r,c}
\le N_c^{\mathrm{core}} .
\end{aligned}
\end{equation}
For decision window $w$, candidate cross-tier movements are
$\mathcal{M}_w=\{m\mid m=(r_m,s_m,d_m,S_m^{\mathrm{mig}},y_m)\}$
where $y_m$ indicates whether migration $m$ is executed.

\subsection{Communication, Storage, and Pressure}

The communication classes are
$\mathcal{K}_L=\{\mathrm{disp},\mathrm{gath},\mathrm{mem},\mathrm{mig}\}$
A request $h=(s_h,d_h,D_h,\mathcal{P}_h,z_h)$ records source, destination, bytes, route path, and an activation indicator. Dispatch requests are generated by $(s_g,c_r,D_g^{\mathrm{act}},\mathcal{P}(s_g,c_r),x_{g,r})$, gather requests by $(c_r,a_g,D_g^{\mathrm{out}},\mathcal{P}(c_r,a_g),x_{g,r})$, memory requests by $((t_r,c_r,b_r),c_r,D_{g,r}^{\mathrm{wt}},\mathcal{P}_{\mathrm{mem}}(t_r,c_r,b_r),x_{g,r})$, and migration requests by $(s_m,d_m,S_m^{\mathrm{mig}},\mathcal{P}(s_m,d_m),y_m)$.

Physical link load is separated from weighted scheduling pressure. For $\kappa\in\mathcal{K}_L$,
\begin{equation}
\begin{aligned}
L_{u,\kappa}^{\mathrm{phys}}(w)
&=
\sum_{h\in\mathcal{H}_{w,\kappa}}
z_hD_h\mathbb{I}[u\in\mathcal{P}_h],\\
\widetilde{L}_{u,\kappa}(w)
&=\alpha_\kappa L_{u,\kappa}^{\mathrm{phys}}(w),
\end{aligned}
\end{equation}
with $\alpha_\kappa=1$ for foreground dispatch, gather, and memory traffic, and $\alpha_{\mathrm{mig}}$ used only to weight migration pressure. The total physical and weighted loads are
\begin{equation}
\begin{aligned}
L_u^{\mathrm{phys}}(w)&=
\sum_{\kappa\in\mathcal{K}_L}L_{u,\kappa}^{\mathrm{phys}}(w),\\
\widetilde{L}_u(w)&=
\sum_{\kappa\in\mathcal{K}_L}\widetilde{L}_{u,\kappa}(w).
\end{aligned}
\end{equation}
Only $L_u^{\mathrm{phys}}$ is used for physical latency and energy. The route time of request $h$ is
\begin{equation}
T_h^{\mathrm{route}}=
\sum_{u\in\mathcal{P}_h}
\left(\lambda_u+\frac{D_h}{B_u}\right),
\end{equation}
and the network-side critical path of layer $l$ in window $w$ is
\begin{equation}
\begin{aligned}
T_l^{\mathrm{net}}(w)=
\max\Bigg(&
\max_{u\in E_{\mathrm{hw}}}
\frac{L_u^{\mathrm{phys}}(w)}{B_u},\\
&
\max_{\kappa\in\mathcal{K}_L}
\max_{h\in\mathcal{H}_{w,\kappa}}
z_hT_h^{\mathrm{route}}
\Bigg).
\end{aligned}
\end{equation}

Storage requests are divided into foreground weight access and migration access, $\mathcal{K}_Q=\{\mathrm{fg},\mathrm{mig}\}$. Each access $a=(t_a,c_a,b_a,Q_a,z_a)$ records tier, chiplet, bank, bytes, and activity. With $\beta_{\mathrm{fg}}=1$ and $\beta_{\mathrm{mig}}=\alpha_{\mathrm{mig}}$, bank traffic is
\begin{equation}
\begin{aligned}
Q_{t,c,b}^{\mathrm{phys}}(w)
&=
\sum_{\chi\in\mathcal{K}_Q}
\sum_{a\in\mathcal{A}_{w,\chi}}
z_aQ_a\mathbb{I}[(t_a,c_a,b_a)=(t,c,b)],\\
\widetilde{Q}_{t,c,b}(w)
&=
\sum_{\chi\in\mathcal{K}_Q}\beta_\chi
\sum_{a\in\mathcal{A}_{w,\chi}}
z_aQ_a\mathbb{I}[(t_a,c_a,b_a)=(t,c,b)].
\end{aligned}
\end{equation}
The bank service time is
\begin{equation}
T_{t,c,b}^{\mathrm{bank}}(w)=
\mathbb{I}[Q_{t,c,b}^{\mathrm{phys}}(w)>0]Lat_{t,c,b}
+\frac{Q_{t,c,b}^{\mathrm{phys}}(w)}{BW_{t,c,b}}.
\end{equation}
Thus
\begin{equation}
T_l^{\mathrm{bank}}(w)=
\max_{t\in\mathcal{T}}\max_{c\in\mathcal{C}}
\max_{b\in\mathcal{B}_{t,c}}
T_{t,c,b}^{\mathrm{bank}}(w),
\end{equation}
and memory-side latency is the maximum of bank service and memory-route service:
\begin{equation}
\begin{aligned}
T_l^{\mathrm{mem}}(w)=
\max\Bigg(&T_l^{\mathrm{bank}}(w),\
\max_{u\in E_{\mathrm{hw}}}
\frac{L_{u,\mathrm{mem}}^{\mathrm{phys}}(w)}{B_u},\\
&\max_{h\in\mathcal{H}_{w,\mathrm{mem}}}
z_hT_h^{\mathrm{route}}\Bigg).
\end{aligned}
\end{equation}

\subsection{Latency, Energy Accounting, and Runtime State}

Replica $r$ processes
$N_r=\sum_{g\in\mathcal{G}_l}x_{g,r}N_g$
tokens and performs
$Ops_r=N_rOps_{l,e(r)}^{\mathrm{token}}$
expert operations. For an active replica, compute time is
\begin{equation}
T_r^{\mathrm{comp}}=
\frac{Ops_r}{k_{r,c_r}\Pi_{\mathrm{core}}\eta_r},
\qquad \xi_r=1,
\end{equation}
where $\eta_r\in(0,1]$ is the legal-tile utilization factor. The implemented artifact uses $Ops_{l,e}^{\mathrm{token}}=2d_{\mathrm{model}}d_{\mathrm{ffn}}$ for a two-layer FFN expert and $3d_{\mathrm{model}}d_{\mathrm{ffn}}$ for a SwiGLU expert. The layer compute critical path is
$T_l^{\mathrm{comp}}=\max_{r\in\mathcal{R}_l:\xi_r=1}T_r^{\mathrm{comp}}$

LEGOSim~\cite{lin2025legosim} records each token group's stage times over
$\mathcal{K}_T=\{\mathrm{gate},\mathrm{disp},\mathrm{queue},\mathrm{mem},\mathrm{comp},\mathrm{gath},\mathrm{combine}\}$
The group completion time and layer latency are
\begin{equation}
C_g=\sum_{\theta\in\mathcal{K}_T}T_{g,\theta},
\qquad
T_l^{\mathrm{MoE}}=\max_{g\in\mathcal{G}_l}C_g .
\end{equation}
At the stage level, the same event graph is summarized as
\begin{equation}
\begin{aligned}
T_l^{\mathrm{MoE}}={}&
T_l^{\mathrm{gate}}+T_l^{\mathrm{disp}}+T_l^{\mathrm{queue}}\\
&+\max(T_l^{\mathrm{comp}},T_l^{\mathrm{mem}})
+T_l^{\mathrm{gath}}+T_l^{\mathrm{combine}} .
\end{aligned}
\end{equation}
This stage-level expression summarizes the routed-MoE event graph; it complements event-level simulation. The max term summarizes compute--memory overlap under compatible event dependencies; serialization caused by dependencies or resource contention is captured by the event-level completion time. The \toolname{} control objective and the stage attribution use the LEGOSim group completion time for routed MoE layers.

\textbf{Serving-window completion metric.} For each serving window $w$, LEGOSim records the full event-graph completion time:
\begin{equation}
T_{\mathrm{E2E}}(w)=
\mathrm{Finish}_{\mathrm{full}}(w)-\mathrm{Start}_{\mathrm{full}}(w).
\label{eq:e2e-latency}
\end{equation}
The stage attribution uses routed-MoE events, and $T_{\mathrm{E2E}}(w)$ aggregates all instantiated model events in the serving window.

For evaluation, window energy accounting is decomposed into compute, link, storage, migration-control, and gating terms:
\begin{equation}
\begin{aligned}
E(w)={}&E_{\mathrm{comp}}(w)+E_{\mathrm{NoC}}(w)
+E_{\mathrm{D2D}}(w)\\
&+E_{\mathrm{SRAM}}(w)+E_{\mathrm{HBM}}(w)
+E_{\mathrm{DRAM}}(w)\\
&+E_{\mathrm{mig}}(w)+E_{\mathrm{gate}}(w).
\end{aligned}
\end{equation}
Following RAPL power capping~\cite{petoumenos2015power,haidar2018power}, the configured window energy budget is computed from the system power cap and the decision-window length. When the accumulated window energy exceeds this budget, the following decision window is modeled with reduced effective compute throughput, capturing the DVFS and clock-throttling response used to enforce power limits.
Compute and link energy are
$E_{\mathrm{comp}}(w)=\sum_{r\in\mathcal{R}_l}Ops_r\epsilon_{\mathrm{MAC}}$
$E_{\kappa}(w)=\sum_{u\in\mathcal{U}_{\kappa}}L_u^{\mathrm{phys}}(w)\epsilon_u,\ \kappa\in\{\mathrm{NoC},\mathrm{D2D}\}$
Storage energy uses physical bank traffic:
$E_t(w)=\sum_{c\in\mathcal{C}}\sum_{b\in\mathcal{B}_{t,c}}Q_{t,c,b}^{\mathrm{phys}}(w)\epsilon_{t,c,b}$
Migration data movement is already charged through link and bank traffic, so the migration term covers only directory and control overhead:
\begin{equation}
E_{\mathrm{mig}}(w)=
\sum_{m\in\mathcal{M}_w}y_mE_m^{\mathrm{ctrl}},
\qquad
E_{\mathrm{gate}}(w)=
\sum_{g\in\mathcal{G}_l}E_g^{\mathrm{gate}}.
\end{equation}

The runtime pressure state consumed by the mapper is
$P_w=(P_w^{\mathrm{link}},P_w^{\mathrm{bank}},P_w^{\mathrm{tier}},P_w^{\mathrm{load}},P_w^{\mathrm{mig}})$
The implemented pressure counters include
\begin{equation}
\begin{aligned}
P_w^{\mathrm{load}}
&=
\max_{c\in\mathcal{C}}
\frac{
\sum_{r\in\mathcal{R}_l:c_r=c,\xi_r=1}
\widehat{T}^{\mathrm{comp}}_r(w)}
{\Delta_w},\\
P_w^{\mathrm{link}}
&=
\max_{u\in E_{\mathrm{hw}}}
\frac{\widetilde{L}_u(w)}{B_u\Delta_w},\\
P_w^{\mathrm{bank}}
&=
\max_{t,c,b}
\frac{\widetilde{Q}_{t,c,b}(w)}
{BW_{t,c,b}\Delta_w}.
\end{aligned}
\end{equation}
Here $\Delta_w$ is the decision-window length, and $\widehat{T}^{\mathrm{comp}}_r(w)$ is the short-window compute service-time estimate for replica $r$ under routed demand. Thus $P_w^{\mathrm{load}}$ measures service-time utilization rather than static core occupancy. $P_w^{\mathrm{tier}}$ records tier-capacity scarcity and $P_w^{\mathrm{mig}}$ records exposed migration pressure. These state variables connect the physical substrate in Figure~\ref{fig:hardware-substrate} to the latency-minimization formulation in Section~\ref{sec:formulation}; they are not optimized as independent objectives.

\section{Problem Formulation}
\label{sec:formulation}

Based on the \threed{} MoE system model in Section~\ref{sec:system}, we define pressure-aware hot-expert residency mapping as an online constrained optimization problem. In each decision window $w$, the system observes the routed token groups, current resident expert replicas, tier capacities, and runtime link and bank pressure. The goal is to choose the resident replica set, physical placement, and token-to-replica assignment that minimize routed-MoE completion cost while satisfying routing correctness and hardware resource constraints.

\subsection{Decision Variables}

For decision window $w$, the runtime mapping decision is
$M_w=(\mathcal{R}_w,\phi_w,x_w)$
Here $\mathcal{R}_w$ is the set of expert replicas retained in the current window. The placement function
$\phi_w(r)=(t_r,c_r,b_r)$
maps replica $r$ to storage tier $t_r$, execution chiplet $c_r$, and backing memory bank $b_r$. The token-assignment variable $x_{g,r}\in\{0,1\}$ indicates whether routed token group $g$ executes on replica $r$.

Thus, $M_w$ captures three coupled runtime decisions: which expert replicas remain resident, where those replicas are placed across tiers and chiplets, and which legal replica receives each routed token group.

\subsection{Latency Objective}

\textbf{Latency objective.} Given mapping $M_w$, \toolname{} minimizes routed MoE-FFN completion cost:
\begin{equation}
\min_{M_w}\ \sum_{l\in\mathcal{L}} T_l^{\mathrm{MoE}}(M_w).
\label{eq:latency-objective}
\end{equation}
Here $\mathcal{L}$ is the set of modeled MoE layers, and $T_l^{\mathrm{MoE}}(M_w)$ is the routed MoE execution latency of layer $l$ under mapping $M_w$. The end-to-end serving-window metric $T_{\mathrm{E2E}}$ is defined in Equation~\ref{eq:e2e-latency}.

\dtd{} traffic, NoP congestion, memory-bank pressure, compute imbalance, tier-capacity scarcity, and migration overhead are factors that change $T_l^{\mathrm{MoE}}(M_w)$ through dispatch, queueing, compute, memory-access, gather, and migration exposure. Communication, storage, and replica movement are therefore not final goals by themselves; the control goal is to reduce routed MoE execution latency produced by their interaction, thereby reducing serving-window completion time.

\subsection{Constraints}

First, the mapping must satisfy routing correctness. Each token group is assigned to exactly one legal replica of its target expert:
\begin{equation}
\sum_{r\in\mathcal{R}_{l,e_g}}x_{g,r}=1,
\qquad
\forall l\in\mathcal{L},\ g\in\mathcal{G}_l.
\label{eq:form-routing}
\end{equation}
Here $e_g$ is the target expert selected for token group $g$, and $\mathcal{R}_{l,e_g}$ is the resident replica set of that expert in layer $l$. This constraint guarantees that each token group is executed once and only by a legal replica of its routed expert.

Second, the mapping must satisfy bank-capacity constraints. For any storage tier, chiplet, and bank, resident expert weights, activation/token-buffer reservation, and metadata overhead cannot exceed bank capacity:
\begin{equation}
\begin{aligned}
&\sum_{r:\phi_w(r)=(t,c,b)}
S_{\ell(r),e(r)}^{\mathrm{wt}}
+A_{t,c,b}+M_{t,c,b} \\
&\hspace{3.0em}\le Cap_{t,c,b},
\qquad
\forall t\in\mathcal{T},\ c\in\mathcal{C},\ b\in\mathcal{B}_{t,c}.
\end{aligned}
\label{eq:form-capacity}
\end{equation}
Here $S_{\ell(r),e(r)}^{\mathrm{wt}}$ is the weight size of replica $r$'s logical expert, $A_{t,c,b}$ reserves activation and token-buffer space, $M_{t,c,b}$ accounts for directory, version, and replica-management metadata, and $Cap_{t,c,b}$ is the capacity of bank $(t,c,b)$.

Finally, the mapping must satisfy replica-count and total residency-budget constraints. Each expert keeps at least one legal replica and no more than the allowed maximum:
\begin{equation}
1\le |\mathcal{R}_{l,e}|\le R_{l,e}^{\max},
\qquad
\forall l\in\mathcal{L},\ e\in\mathcal{E}_l.
\label{eq:form-replica-count}
\end{equation}
The total resident expert-weight footprint cannot exceed the system residency budget:
\begin{equation}
\sum_{l\in\mathcal{L}}
\sum_{e\in\mathcal{E}_l}
S_{l,e}^{\mathrm{wt}}|\mathcal{R}_{l,e}|
\le B_{\mathrm{res}}.
\label{eq:form-res-budget}
\end{equation}
Here $R_{l,e}^{\max}$ is the maximum allowed replica count for expert $e$ in layer $l$, and $B_{\mathrm{res}}$ is the total expert residency budget.

In summary, the problem is to choose expert replicas, their physical tier/chiplet/bank locations, and token-group assignments under finite SRAM, HBM, and DRAM capacity so that the routed-MoE contribution to full-model latency is minimized. The core challenge is not simply reducing communication or unconditionally increasing hot-expert replicas; it is finding a runtime mapping that lowers final execution latency under coupled capacity, link, bank, and compute-load constraints.

\section{Method Design}
\label{sec:method}

\subsection{Overview: Challenge-Oriented Two-Timescale Control}

Section~\ref{sec:formulation} defines pressure-aware hot-expert residency mapping as an online constrained optimization problem whose objective is to reduce full-model latency by minimizing the policy-controlled routed-MoE component. The runtime decision is $M_w=(\mathcal{R}_w,\phi_w,x_w)$, where $\mathcal{R}_w$ is the expert-replica set in the current window, $\phi_w$ places each replica across memory tier, chiplet, and bank, and $x_w$ maps routed token groups to expert replicas.

Solving this problem directly is impractical. First, SRAM, HBM, and DRAM in a \threed{} system have different capacities, access paths, and sharing scopes, so expert-replica placement simultaneously changes dispatch latency, queueing latency, memory-access latency, and gather latency. Second, hot-expert replicas have cross-window benefits and costs: a new replica may reduce queueing and repeated weight streaming over several future windows, but it immediately consumes capacity and introduces migration traffic and bank pressure. Fixed thresholds or single-window heuristics therefore cannot reliably decide whether a residency update is worth performing.

\toolname{} uses a hybrid two-timescale control framework. A slow residency controller uses a masked feasibility-constrained Double DQN to learn cross-window promotion, retention, demotion, and eviction policies for expert replicas, thereby updating $\mathcal{R}_w$ and $\phi_w$. The action mask enforces replica-count, residency-budget, bank-capacity, and migration-path feasibility before action selection, so the controller chooses only from legal residency edits. A fast token mapper uses an explicit normalized pressure proxy under the current expert-replica directory to select the execution replica for each token group, thereby updating $x_w$. Together, these two loops construct the runtime mapping $M_w$ used in Section~\ref{sec:formulation}.

This decomposition corresponds to the two challenges in Section~\ref{sec:motivation}. For Challenge 1, \toolname{} no longer places experts by communication distance alone. The residency controller observes tier capacity, bank pressure, link pressure, and compute load, and then combines these signals with latency-aware placement to select multi-level resident locations. For Challenge 2, \toolname{} does not treat expert replicas as redundancy that should be increased unconditionally. Instead, the masked Double-DQN controller learns the long-term tradeoff between replica benefit and migration cost. Once a replica layout is chosen, the fast token mapper lets tokens avoid congested queues, congested links, and pressured memory banks in real time.

\subsection{Runtime State and Expert Demand}

In each decision window $w$, \toolname{} first extracts per-expert demand from the routing result. For expert $e$ in layer $l$, the window token demand is
\begin{equation}
d_{l,e}(w)=
\sum_{g\in\mathcal{G}_l(w):e_g=e} N_g .
\end{equation}
Here $\mathcal{G}_l(w)$ is the routed token-group set of layer $l$ in window $w$, $e_g$ is the target expert of token group $g$, and $N_g$ is the number of tokens in that group.

Because expert hotness changes with serving mode, batch composition, and request domain, \toolname{} triggers migration from smoothed hotness using an exponentially weighted moving average (EWMA):
$\bar d_{l,e}(w)=\alpha \bar d_{l,e}(w-1)+(1-\alpha)d_{l,e}(w)$
where $\alpha\in[0,1]$ controls the balance between historical and current windows. Smoothed hotness separates persistent hot experts from short burst experts, avoiding frequent expert-weight migration caused by one-window traffic fluctuation.

To represent physical pressure in the \threed{} system, \toolname{} uses the runtime pressure state defined in Section~\ref{sec:system}:
$P_w=(P_w^{\mathrm{load}},P_w^{\mathrm{link}},P_w^{\mathrm{bank}},P_w^{\mathrm{tier}},P_w^{\mathrm{mig}})$
Here $P_w^{\mathrm{load}}$ captures compute-load pressure, $P_w^{\mathrm{link}}$ captures NoP and \dtd{} link pressure, $P_w^{\mathrm{bank}}$ captures memory-bank pressure, $P_w^{\mathrm{tier}}$ captures tier-capacity pressure, and $P_w^{\mathrm{mig}}$ captures exposed migration pressure.

The slow controller observes the following state in window $w$:
\begin{equation}
s_w =
\big[
\bar d_{l,e}(w),\sigma_{l,e}(w),H_{l,e}^{\mathrm{src}}(w),
\mathcal{D}_{w-1},P_w,Age_w,B_w
\big].
\end{equation}
Here $\sigma_{l,e}(w)$ is the recent fluctuation of expert demand, $H_{l,e}^{\mathrm{src}}(w)$ is the entropy of the expert's request sources across chiplets, $\mathcal{D}_{w-1}$ is the previous-window expert-to-replica directory, $Age_w$ is the current replica residency age, and $B_w$ is the remaining capacity of each tier and bank. The state jointly captures expert hotness, hotness stability, source locality, current replica layout, and hardware pressure, allowing the controller to estimate how a residency update may affect future latency.

\subsection{Masked Double DQN for Slow Residency Updates}

\toolname{} models slow expert-residency update as a masked MDP with hard feasibility constraints. In each decision window $w$, the controller selects a legal residency edit action $a_w$ from state $s_w$ to update the expert-replica set and part of the physical placement. The action set is
\begin{equation}
\begin{aligned}
a_w\in\{&
\mathrm{no\text{-}op},\
\mathrm{promote}(e,t,c),\
\mathrm{retain}(e,t,c),\\
&\mathrm{demote}(e,t,c),\
\mathrm{evict}(e,t,c)\}.
\end{aligned}
\end{equation}
Here no-op means that the current window performs no residency update; promote moves or creates a replica of expert $e$ in a nearer tier $t$ and chiplet $c$; retain keeps an existing replica; demote moves a replica to a tier with more capacity or lower pressure; and evict removes an extra replica that no longer provides latency benefit.

To control action-space size, \toolname{} does not enumerate all experts and all locations. It generates actions only for a candidate expert set $\mathcal{H}_w$, including persistent hot experts, warm experts that overflow SRAM, and stale replicas that have not been useful for a long time. For each candidate expert, the target tier/chiplet location must belong to a feasible candidate set $\Omega_e(w)$ that satisfies capacity and path constraints.

Before action execution, \toolname{} applies a legal mask to remove all actions that violate constraints. Legal actions must satisfy
\begin{equation}
1\le |\mathcal{R}_{l,e}|\le R_{l,e}^{\max},
\end{equation}
\begin{equation}
\sum_{l\in\mathcal{L}}
\sum_{e\in\mathcal{E}_l}
S_{l,e}^{\mathrm{wt}}|\mathcal{R}_{l,e}|
\le B_{\mathrm{res}},
\end{equation}
\begin{equation}
\begin{aligned}
&\sum_{r:\phi(r)=(t,c,b)}
S_{\ell(r),e(r)}^{\mathrm{wt}}
+A_{t,c,b}+M_{t,c,b}\\
&\hspace{3.0em}\le Cap_{t,c,b}.
\end{aligned}
\end{equation}
Thus, an action cannot evict the last legal replica of an expert, exceed the total residency budget, exceed any bank capacity, or force a cross-tier migration when the path is unavailable. With legal masking, the controller chooses directly inside the feasible action space, keeping action selection feasible before reward evaluation.

\subsection{Reward: Same-Window Counterfactual Benefit}

Residency actions are trained with a same-window counterfactual reward. Adjacent windows can differ in token demand, so \toolname{} evaluates the selected action and the previous layout on the same routed token groups. This assigns credit to residency and token-to-replica decisions without mixing in workload drift across windows.

For each training window $w$, the simulator evaluates the same routed token groups $\mathcal{G}_w$ under two layouts. The counterfactual layout keeps the previous directory and applies only the fast token mapper:
\begin{equation}
T^{\mathrm{cf}}_w =
T^{\mathrm{MoE}}(\mathcal{G}_w,\mathcal{D}_{w-1},
\mathrm{FastTokenMap}).
\end{equation}

The acted layout schedules the selected residency action $a_w$, inserts the corresponding migration events, publishes only completed replicas, and then applies the same fast token mapper to the published directory:
\begin{equation}
\begin{aligned}
T^{\mathrm{act}}_w ={}&
T^{\mathrm{MoE}}(\mathcal{G}_w,
\mathcal{D}^{\mathrm{act}}_w,
\mathrm{FastTokenMap}),\\
(\mathrm{Pnd}^{\mathrm{act}}_w,\mathrm{MigEvents}^{\mathrm{act}}_w) ={}&
\mathrm{ScheduleAction}(a_w,\mathcal{D}_{w-1}),\\
\mathcal{D}^{\mathrm{act}}_w ={}&
\mathrm{PublishCompleted}(\mathrm{Pnd}^{\mathrm{act}}_w,\mathcal{D}_{w-1}).
\end{aligned}
\end{equation}

The reward is
\begin{equation}
R_w =
\frac{T^{\mathrm{cf}}_w - T^{\mathrm{act}}_w}{T^{\mathrm{cf}}_w}
-\mu_1\widehat{T}^{\mathrm{mig}}_w
-\mu_2 C^{\mathrm{churn}}_w .
\end{equation}

Here $T^{\mathrm{cf}}_w$ and $T^{\mathrm{act}}_w$ are measured on the same routed-token demand, so the first term attributes latency change to the residency edit rather than to workload drift. $\widehat{T}^{\mathrm{mig}}_w$ penalizes foreground-exposed migration time, and $C^{\mathrm{churn}}_w$ penalizes frequent promotion, demotion, or eviction. The counterfactual evaluation is used only during offline training. During evaluation, the controller is frozen and does not execute both layouts online.

During offline training, \toolname{} uses Masked Double DQN. For transition $(s_w,a_w,R_w,s_{w+1})$, the target value is
\begin{equation}
y_w =
R_w+\gamma
Q_{\bar{\theta}}
\left(
s_{w+1},
\arg\max_{a'\in\mathcal{A}_{w+1}^{\mathrm{legal}}}
Q_{\theta}(s_{w+1},a')
\right),
\end{equation}
and the training loss is
$\mathcal{L}(\theta)=(Q_{\theta}(s_w,a_w)-y_w)^2$.
Here $\gamma$ is the future-reward discount factor, $\mathcal{A}_{w+1}^{\mathrm{legal}}$ is the next-window action set after legal masking, $Q_{\theta}$ is the online network, and $Q_{\bar{\theta}}$ is the target network. Double DQN reduces overestimation in action-value estimates, while the legal mask ensures that training and inference always satisfy capacity, replica-count, and migration-feasibility constraints.

\subsection{Latency-Aware Multi-Level Placement}

Challenge 1 arises because \threed{} systems provide multiple memory tiers with different resource properties. SRAM has the lowest latency but the tightest capacity. HBM provides larger intermediate capacity but is limited by bank bandwidth. DRAM has the largest capacity but introduces shared-I/O and \dtd{}/NoP pressure. Therefore, placing an expert nearest to the token source does not guarantee the lowest execution latency.

\toolname{} addresses this issue with the RL residency controller and a latency-aware placement rule. The RL controller decides which residency tier an expert should occupy according to expert hotness, hotness stability, and system pressure. Persistent hot experts compete for near SRAM residency regions, warm experts are preferentially kept in HBM to avoid repeated streaming from shared DRAM, and cold experts retain one legal replica in the shared capacity tier.

When the RL controller selects a promote or retain action, \toolname{} chooses the concrete tier/chiplet/bank location from legal candidates:
\begin{equation}
(t_r,c_r,b_r)=
\arg\min_{(t,c,b)\in\Omega_r}
C_{\mathrm{place}}(r,t,c,b),
\end{equation}
where $\Omega_r$ is the candidate set satisfying capacity and path feasibility. The placement cost is
\begin{equation}
\begin{aligned}
C_{\mathrm{place}}(r,t,c,b)={}&
\alpha_p D_{\mathrm{src}}(e(r),c)
+\beta_p Q_c
+\chi_p U_{t,c,b}^{\mathrm{bank}}\\
&+\psi_p U_t^{\mathrm{cap}}
-\eta_p D_{\mathrm{div}}(r,c).
\end{aligned}
\end{equation}
Here $D_{\mathrm{src}}(e(r),c)$ is the weighted distance from the routed-token sources of expert $e(r)$ to chiplet $c$, $Q_c$ is the current compute-queue pressure on chiplet $c$, $U_{t,c,b}^{\mathrm{bank}}$ is the access pressure of bank $(t,c,b)$, $U_t^{\mathrm{cap}}$ is the capacity scarcity of tier $t$, and $D_{\mathrm{div}}(r,c)$ is the location diversity between this replica and existing replicas of the same expert.

This placement rule does not minimize communication distance alone. It prevents hot experts from being concentrated on the same source-near chiplet or the same pressured bank. By jointly considering source locality, compute queue, bank pressure, tier capacity, and replica diversity, \toolname{} turns expert-weight placement into a local approximation of its effect on $T_l^{\mathrm{MoE}}$, reducing latency uncertainty caused by heterogeneous tiers and coupled physical pressure.

\subsection{Bounded RL-Guided Replica Control}

Challenge 2 arises because expert replicas can reduce hot-expert queueing, but they also consume SRAM/HBM capacity and introduce weight migration, bank access, and link occupancy. When capacity is insufficient or bank pressure is high, a fixed replication policy can re-concentrate load and even increase latency.

\toolname{} treats expert replicas as constrained residency resources with explicit capacity and migration costs. The RL controller learns whether an action can amortize its migration cost across multiple windows. For a hot expert, the controller promotes a new replica only when the replica provides positive long-term latency benefit. For a short burst expert, the controller tends to choose no-op or retain to avoid migration churn. If a replica no longer contributes to load spreading or repeated-streaming reduction, the controller demotes or evicts it to release near-tier capacity.

To prevent uncontrolled replication, \toolname{} bounds the number of residency updates per window:
$\sum_{m\in\mathcal{M}_w} y_m \le U^{\max}$
where $\mathcal{M}_w$ is the candidate migration or residency-update set in window $w$, $y_m$ indicates whether update $m$ is executed, and $U^{\max}$ is the maximum number of updates allowed in one window.

\toolname{} also maintains a residency age for each replica. A newly promoted replica can be evicted only after it reaches a minimum residency time:
$\mathrm{age}(r)\ge A^{\min}$
This mechanism prevents replicas from oscillating among hot, warm, and cold states. Compared with fixed replication, \toolname{}'s replica control is value-guided and bounded: the masked Double DQN controller selects one legal residency action per window, spending residency capacity only when the learned long-term value justifies the update and releasing resources when a replica is no longer useful.

\subsection{Fast Token-to-Replica Mapping}

The slow RL controller decides which replicas exist and where they are placed. Inside each window, however, the runtime must still decide which replica receives each token group. This decision is high-frequency, latency-sensitive, and must strictly preserve routing correctness. \toolname{} therefore uses an explicit normalized pressure proxy for deterministic selection of $x_w$.

For token group $g$ with target expert $e_g$, \toolname{} chooses the lowest-cost replica from the legal replica set $\mathcal{R}_{l,e_g}$:
\begin{equation}
r_g^\ast=
\arg\min_{r\in\mathcal{R}_{l,e_g}}
C_{\mathrm{tok}}(g,r),
\end{equation}
and sets
\begin{equation}
x_{g,r}=
\begin{cases}
1, & r=r_g^\ast,\\
0, & \text{otherwise}.
\end{cases}
\end{equation}
The token-assignment cost is a normalized pressure proxy rather than a physical latency equation:
\begin{equation}
C_{\mathrm{tok}}(g,r)=
\omega_q C^{\mathrm{queue}}_{g,r}
+\omega_p C^{\mathrm{path}}_{g,r}
+\omega_b C^{\mathrm{bank}}_{g,r}
+\omega_s C^{\mathrm{stream}}_{g,r}.
\end{equation}
Here $C^{\mathrm{queue}}_{g,r}$ measures queued service at replica $r$, $C^{\mathrm{path}}_{g,r}$ measures dispatch and gather path pressure, $C^{\mathrm{bank}}_{g,r}$ measures backing-bank pressure, and $C^{\mathrm{stream}}_{g,r}$ measures tier-dependent streaming cost. The weights are fixed during evaluation and only rank legal replicas inside the current directory; physical latency is still measured by the event-level simulator.

In execution, \toolname{} assigns tokens in small token blocks and updates replica load, chiplet load, link load, and bank pressure after each block. This avoids sending all tokens to the same replica using stale window-start state. Fast token mapping allows \toolname{} to exploit existing replicas to spread hot-expert queues and avoid congested links or banks without changing the residency layout.

\subsection{Stable Migration and Versioned Directory}

Promotion, demotion, and eviction are not zero-cost operations. Cross-tier migration consumes \dtd{}/NoP paths, HBM interfaces, and bank service time. If migration conflicts with foreground dispatch/gather traffic, it can become a new latency bottleneck.

\toolname{} therefore applies migration throttling and a versioned directory when executing RL residency actions. For a candidate migration path $\mathcal{P}_{\mathrm{mig}}$, migration is allowed only when foreground utilization plus migration utilization remains below a threshold on every path link:
\begin{equation}
U_u^{\mathrm{fg}}(w)+U_u^{\mathrm{mig}}(w)\le \eta_u,
\qquad
\forall u\in\mathcal{P}_{\mathrm{mig}} .
\end{equation}
Here $U_u^{\mathrm{fg}}(w)$ is the utilization of foreground dispatch, gather, and memory traffic on link $u$, $U_u^{\mathrm{mig}}(w)$ is migration-traffic utilization, and $\eta_u$ is the safety threshold used to avoid interfering with foreground execution. If this condition is not satisfied, the migration is delayed and the current window falls back to no-op or retain.

For directory consistency, \toolname{} uses a versioned expert-to-replica directory. When a replica is promoted or demoted, the system does not overwrite the old replica. Instead, it copies weights in chunks into a pending region in the target tier. A new directory version is published only after all chunks complete and the target location satisfies capacity and pressure constraints. Token groups already assigned to the old version finish on the old location, while later token groups use the new directory. FastTokenMap always consumes the last published directory. A pending replica created by a residency action is not a legal execution target until all migration chunks have completed and the new directory version has been published. Therefore, migration consumes physical link and bank resources before it can produce token-mapping benefit. Because expert weights are read-only during inference, this mechanism prevents partial-copy reads without requiring a complex coherence protocol.

\subsection{Runtime Flow}

Algorithm~\ref{alg:hcrmap_online} gives the evaluation-time pseudocode of the \toolname{} slow residency control loop. The slow controller does not greedily rank residency edits by one-window estimated benefit. Instead, it uses a frozen masked Double DQN checkpoint to select one legal residency action per decision window. The legal action mask enforces replica-count, residency-budget, bank-capacity, and migration-feasibility constraints before action selection. Algorithm~\ref{alg:fast_token_map} details the deterministic fast token mapping used after the current published directory is fixed.

\begin{algorithm}[t]
\caption{HCRMap Online Residency Control with a Frozen Masked Double DQN}
\label{alg:hcrmap_online}
\footnotesize
\KwIn{Routed token groups $\mathcal{G}_w$; previous directory $\mathcal{D}_{w-1}$; previous replicas $\mathcal{R}_{w-1}$; previous placement $\phi_{w-1}$; frozen Q-network $Q_\theta$.}
\KwOut{Current directory $\mathcal{D}_w$; replicas $\mathcal{R}_w$; placement $\phi_w$; token assignment $x_w$.}

\For{each decision window $w$}{
    Compute expert demand $d_{l,e}(w) \leftarrow \sum_{g\in\mathcal{G}_l(w):e_g=e}N_g$\;
    Update smoothed demand $\bar{d}_{l,e}(w)\leftarrow\alpha\bar{d}_{l,e}(w-1)+(1-\alpha)d_{l,e}(w)$\;
    Collect pressure state $P_w\leftarrow(P^{\mathrm{load}}_w,P^{\mathrm{link}}_w,P^{\mathrm{bank}}_w,P^{\mathrm{tier}}_w,P^{\mathrm{mig}}_w)$\;
    Build state $s_w\leftarrow[\bar{d}_{l,e}(w),\sigma_{l,e}(w),H^{\mathrm{src}}_{l,e}(w),\mathcal{D}_{w-1},P_w,Age_w,B_w]$\;

    Generate candidate expert set $\mathcal{H}_w$ from persistent hot experts, SRAM-overflow warm experts, and stale replicas\;
    Initialize action set $\mathcal{A}_w\leftarrow\{\mathrm{no\mbox{-}op}\}$\;
    \For{each $e\in\mathcal{H}_w$ and each $(t,c)\in\Omega_e(w)$}{
        Add $\mathrm{promote}(e,t,c)$, $\mathrm{retain}(e,t,c)$, $\mathrm{demote}(e,t,c)$, and $\mathrm{evict}(e,t,c)$ to $\mathcal{A}_w$\;
    }
    $\mathcal{A}^{\mathrm{legal}}_w\leftarrow\{a\in\mathcal{A}_w\mid C_{\mathrm{rep}}(a)\wedge C_{\mathrm{budget}}(a)\wedge C_{\mathrm{bank}}(a)\wedge C_{\mathrm{mig}}(a)\}$\;
    $a_w\leftarrow\arg\max_{a\in\mathcal{A}^{\mathrm{legal}}_w}Q_\theta(s_w,a)$\;
    $(\mathrm{Pnd}_w,\mathrm{MigEvents}_w)\leftarrow\mathrm{ScheduleAction}(a_w,\mathcal{R}_{w-1},\phi_{w-1},\mathcal{D}_{w-1})$\;
    Insert $\mathrm{MigEvents}_w$ into the LEGOSim event graph subject to migration throttling\;
    $(\mathcal{R}_w,\phi_w,\mathcal{D}_w)\leftarrow\mathrm{PublishCompleted}(\mathrm{Pnd}_w,\mathcal{R}_{w-1},\phi_{w-1},\mathcal{D}_{w-1})$\;
    $x_w\leftarrow\mathrm{FastTokenMap}(\mathcal{G}_w,\mathcal{D}_w,P_w)$\;
}
\end{algorithm}

Algorithm~\ref{alg:hcrmap_online} summarizes slow residency control with a frozen Q-network checkpoint. Offline training uses simulator-generated transitions and the same-window counterfactual reward above.

\begin{algorithm}[t]
\caption{Fast Token Mapping under the Current Residency Layout}
\label{alg:fast_token_map}
\KwIn{Routed token groups $\mathcal{G}_w$; current directory $\mathcal{D}_w$; runtime pressure state $P_w$.}
\KwOut{Token-to-replica assignment $x_w$.}
\For{each routed token group $g\in\mathcal{G}_w$}{
    Get the target expert $e_g$ and legal replicas $\mathcal{R}_{l,e_g}$ from $\mathcal{D}_w$\;
    \For{each legal replica $r\in\mathcal{R}_{l,e_g}$}{
        Estimate queue pressure $C^{\mathrm{queue}}_{g,r}$\;
        Estimate dispatch/gather path pressure $C^{\mathrm{path}}_{g,r}$\;
        Estimate backing-bank pressure $C^{\mathrm{bank}}_{g,r}$\;
        Estimate tier-dependent streaming cost $C^{\mathrm{stream}}_{g,r}$\;
        Compute pressure-proxy cost
        $\begin{aligned}[t]
        C_{\mathrm{tok}}(g,r)={}&\omega_q C^{\mathrm{queue}}_{g,r}
        +\omega_p C^{\mathrm{path}}_{g,r}\\[-1pt]
        &+\omega_b C^{\mathrm{bank}}_{g,r}
        +\omega_s C^{\mathrm{stream}}_{g,r}.
        \end{aligned}$\;
    }
    Assign $g$ to the lowest-cost replica
    $\begin{aligned}[t]
    r^*={}&\arg\min_{r\in\mathcal{R}_{l,e_g}}\\[-1pt]
    &C_{\mathrm{tok}}(g,r).
    \end{aligned}$\;
    Set $x_{g,r^*}=1$ and $x_{g,r}=0$ for all $r\neq r^*$\;
    Update temporary queue, link, and bank pressure counters\;
}
\Return{$x_w$}\;
\end{algorithm}

Through this flow, \toolname{} turns hot-expert management in a \threed{} system from fixed replication or pure communication optimization into feasibility-masked cross-window residency control. The slow controller learns the long-term tradeoff among capacity, migration, and future latency benefit, while the deterministic fast token mapper keeps per-window replica selection explainable, low-overhead, and routing-correct. Multi-level residency, bounded replica control, and real-time token assignment together reduce the routed-MoE critical path inside full-model execution.

The controller is trained offline with simulator-generated transitions and the same-window counterfactual reward above. Evaluation uses a frozen masked Double-DQN checkpoint. The legal action mask enforces replica-count, residency-budget, bank-capacity, and migration-path constraints, and the runtime applies at most one legal residency update per decision window.

\vspace{-0.5\baselineskip}
\section{Experimental Evaluation}
\label{sec:experiments}
\label{sec:results}

\subsection{Experimental Setup}
\label{sec:eval-setup}

We evaluate \toolname{} using LEGOSim~\cite{lin2025legosim}, a heterogeneous multi-chiplet event-level simulator. \toolname{} and all compared baselines run on the same \threed{} substrate. Each benchmark in Table~\ref{tab:evaluation_setup} instantiates an MoE model configuration with layer topology, number of MoE layers, expert count, Top-$k$, hidden and intermediate dimensions, routing output, and prefill and decode execution shape. Router outputs are collected from model execution and replayed without synthetic popularity generation or skew calibration. The MoE-to-LEGOSim adapter consumes the model-specific routed expert IDs and token-group sizes, and expands each MoE layer into dispatch, queueing, memory access, expert compute, gather, and combine events. All policies use identical per-model routing outputs under the same hardware graph, memory hierarchy, link bandwidths, bank bandwidths, and energy coefficients.

\begin{table}[t]
\centering
\caption{Simulation and hardware configuration.}
\label{tab:sim_hw_config}
\compacttablesetup
\begin{tabular}{@{}P{0.30\columnwidth}P{0.64\columnwidth}@{}}
\toprule
\multicolumn{2}{@{}l}{\textit{Simulation}} \\
\midrule
Simulator & LEGOSim~\cite{lin2025legosim} \\
Model frontend & MoE-to-LEGOSim adapter \\
Benchmark unit & Complete MoE model configuration \\
Router output & Full-model recorded trace \\
\midrule
\multicolumn{2}{@{}l}{\textit{Hardware}} \\
\midrule
Package & 3.5D, UCIe-class D2D~\cite{UCIeConsortium2023_HotChipsTutorial,UCIeConsortium_Specifications} \\
Compute chiplets & 16 \\
IO/memory chiplets & 4 \\
Total cores & 256 \\
Cores/chiplet & 16 \\
Topology & 4$\times$4 mesh \\
Frequency & 1 GHz \\
\midrule
\multicolumn{2}{@{}l}{\textit{Interconnect model}} \\
\midrule
Flit size & 128 bits \\
D2D bandwidth & 256 GB/s per direction; 512 GB/s full-duplex aggregate~\cite{UCIeConsortium2023_HotChipsTutorial,UCIeConsortium_Specifications} \\
Intra-chiplet latency & Router: 2 cycles; link: 1 cycle \\
Inter-chiplet latency & Router: 2 cycles; UCIe-class link target~\cite{UCIeConsortium2023_HotChipsTutorial,UCIeConsortium_Specifications} \\
Routing algorithm & XY routing \\
Buffer depth & 4 flits/router \\
\midrule
\multicolumn{2}{@{}l}{\textit{Memory and execution}} \\
\midrule
Expert footprint & Model-specific expert-weight size \\
SRAM tier & 64 MB stacked near-tier budget; 16 banks; 2.0 TB/s; 1.25 pJ/B~\cite{Wuu2022_3DVCache,Horowitz2014_ComputingEnergy} \\
HBM tier & 8 GB HBM2 subsystem; 32 pseudo-channels; 460 GB/s; 31.8 pJ/B~\cite{AMD2023_AlveoU280_UG1314,OConnor2017_FGDRAM} \\
DRAM tier & 32 GB shared DDR5 tier; 8 DDR5-6400 channels; 409.6 GB/s; 160 pJ/B~\cite{Micron_DDR5_SDRAM_Core,Horowitz2014_ComputingEnergy} \\
Memory budget & $1.3\times$ resident-copy capacity \\
Residency tiers & SRAM hot; HBM warm; DRAM fallback \\
\bottomrule
\end{tabular}
\end{table}

\begin{table}[h]
\centering
\caption{Evaluation setup.}
\label{tab:evaluation_setup}
\compacttablesetup
\begin{tabular}{@{}P{0.30\columnwidth}P{0.64\columnwidth}@{}}
\toprule
Benchmarks & DBRX-MoE (DBRX)~\cite{dbrx2024}; DeepSeek-V2-Lite (DSV2L)~\cite{deepseekv22024}; DeepSeek-V2 (DSV2)~\cite{deepseekv22024}; DeepSeek-V3 (DSV3)~\cite{deepseekv32024}; DeepSeekMoE-16B (DSM16)~\cite{deepseekmoe2024}; Mixtral-8$\times$7B (MX7B)~\cite{jiang2024mixtral}; Mixtral-8$\times$22B (MX22B)~\cite{mixtral8x22b2024}; Qwen1.5-MoE-A2.7B (Qwen)~\cite{qwenmoe2024} \\
\midrule
Prefill & 2048 input tokens \\
Decode & 512 active tokens \\
Compared policies & Hydra; MoEntwine; PIMoE; HCRMap \\
\bottomrule
\end{tabular}
\end{table}

Table~\ref{tab:sim_hw_config} summarizes the simulation and hardware configuration. We evaluate eight MoE models: DeepSeekMoE-16B, DeepSeek-V2-Lite, DeepSeek-V2, DeepSeek-V3, Mixtral-8x7B, Mixtral-8x22B, DBRX, and Qwen1.5-MoE-A2.7B. The two serving settings are prefill inference with 2048 input tokens and decode inference with 512 active sequence tokens.

\textbf{Policy comparison.} Hydra, MoEntwine, PIMoE, and \toolname{} use the same model configuration, recorded router-output trace, and prefill and decode execution shape. Before system mapping, each routed token group is bound to the expert selected by Top-$k$ routing and executes on a legal replica of that expert. The policies manage expert placement, replication, migration, offloading, and token-to-replica assignment under the same \threed{} hardware constraints. Unless otherwise stated, normalized latency uses the single-copy static placement baseline as the denominator within the same model, serving mode, and routing seed. \toolname{} uses the masked feasibility-constrained slow residency controller and latency-aware fast token mapper described in Section~\ref{sec:method}. Energy and \edp{} are tracked alongside latency.

\subsection{Overall Results}
\label{sec:eval-results}

\begin{figure*}[width=\textwidth,pos=t,align=\centering]
\centering
\includegraphics[width=0.97\textwidth]{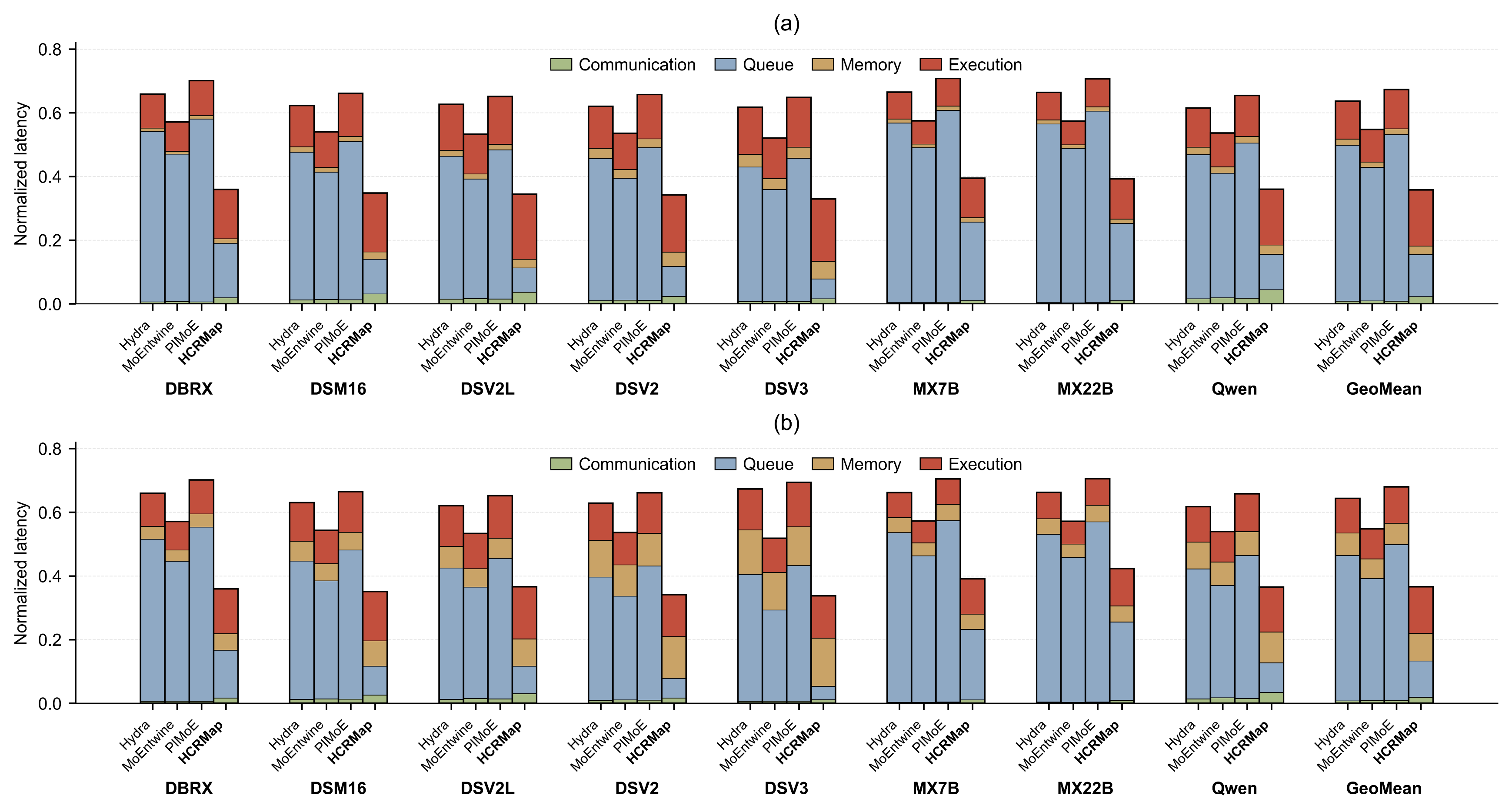}
\caption{Normalized end-to-end latency across eight MoE models: (a) prefill; (b) decode. Stacked bars show stage-level attribution. Lower is better.}
\label{fig:eval-overall-latency}
\end{figure*}

Figure~\ref{fig:eval-overall-latency} reports end-to-end full-model latency for the four expert-management policies across all eight MoE benchmarks. The stacked segments attribute the policy-sensitive portion of the critical path to communication, queueing, memory service, and execution events affected by routed expert management. In the unified LEGOSim simulation, Hydra, MoEntwine, and PIMoE instantiate the same full models on the same \threed{} substrate as \toolname{}. Across eight models, \toolname{} reduces geometric-mean end-to-end latency by 43.6\% and 43.0\% over Hydra, 34.5\% and 33.1\% over MoEntwine, and 46.7\% and 46.0\% over PIMoE in prefill and decode, respectively. The improvement is consistent across benchmarks. The routed-MoE stage attribution also explains the mechanism: Hydra reduces communication traffic but leaves a large queueing component, MoEntwine reduces part of the queueing pressure but still leaves memory and communication service on the critical path, and PIMoE improves memory service for colder work while the hot-expert queue remains visible. \toolname{} is faster because it jointly reduces the dominant queueing and memory-service components through pressure-aware resident mapping and fast token assignment. Bounded additional traffic lowers routed-expert queueing and memory-service contribution on the full-model critical path.

Under matched resident-copy capacity, \toolname{} reduces geometric-mean end-to-end latency by 24.95\% and 23.91\% relative to the strongest fixed-replica baseline in prefill and decode, respectively. When improvement is computed per benchmark first and then averaged, \toolname{} reduces latency by 19.63\%$\pm$2.74\% and 19.04\%$\pm$3.09\%, respectively. The corresponding \edp{} improves by 19.94\%$\pm$2.80\% and 19.86\%$\pm$3.24\%, respectively.

\subsection{Ablation Study}
\label{sec:eval-ablation}

\begin{figure}[pos=!h,align=\centering]
\centering
\ExplSyntaxOn\dim_set:Nn \l_fig_width_dim { \linewidth }\ExplSyntaxOff
\includegraphics[width=0.97\columnwidth,trim=0 20 0 20,clip]{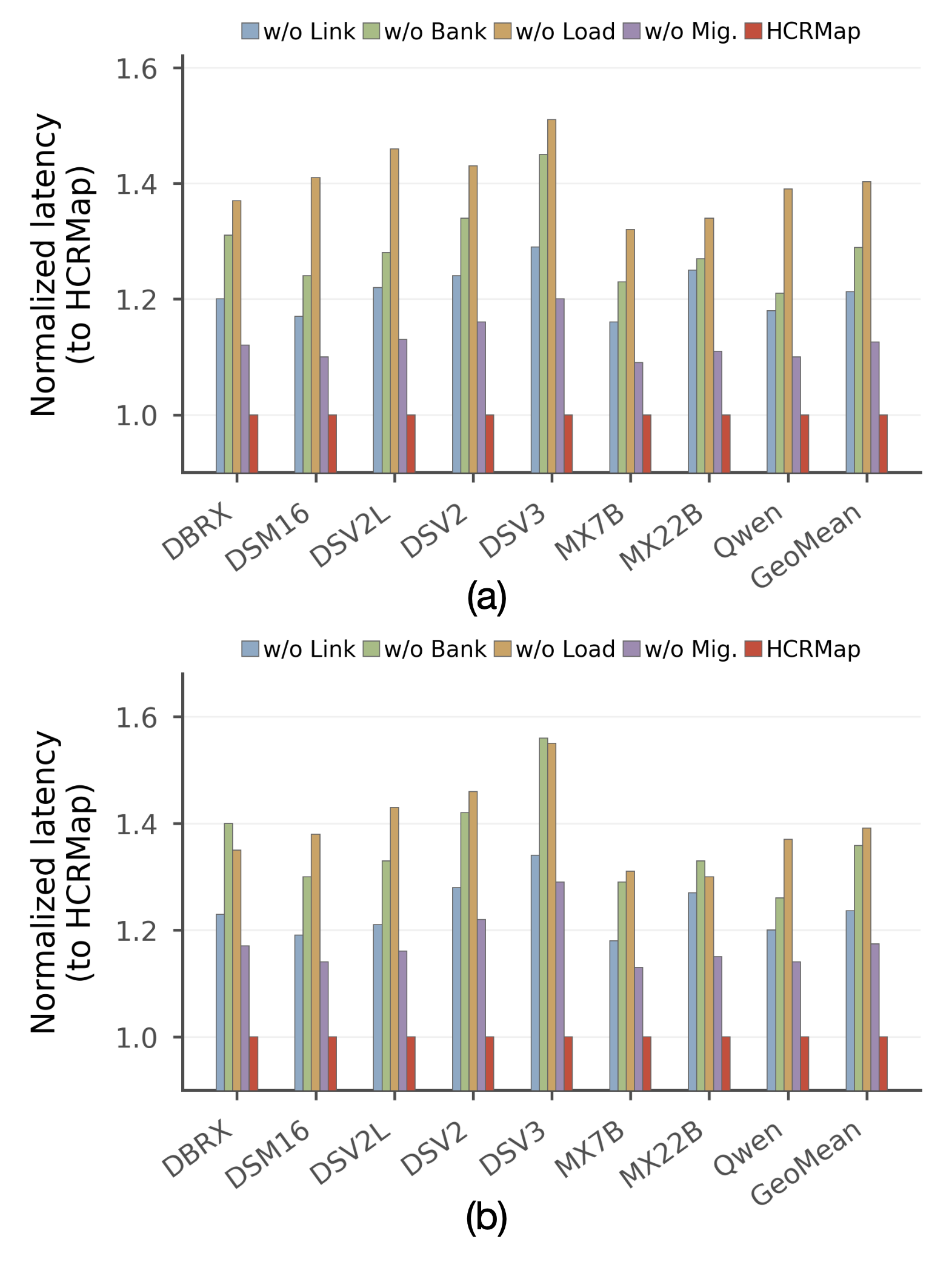}
\vspace{-8pt}
\caption{Pressure-signal ablation normalized to full \toolname{}: (a) prefill; (b) decode. Lower is better.}
\label{fig:eval-ablation}
\end{figure}

Figure~\ref{fig:eval-ablation} isolates the pressure signals used by \toolname{}. Removing link-pressure awareness increases latency because token assignment and migration can reuse already congested package paths. Removing bank-pressure awareness is harmful when hot or warm replicas compete for the same SRAM/HBM banks. Removing compute-load awareness causes routed tokens to concentrate on source-near but already queued replicas. Removing migration-exposure awareness allows useful copies to be created at the wrong time, making migration traffic visible to foreground dispatch, gather, or memory service. The full \toolname{} controller uses all four pressure signals and therefore achieves the lowest normalized end-to-end latency.

We also run selected component checks under matched resident-copy capacity. Compared with static replicas and simple routing, full \toolname{} reduces latency by 24.95\% and 23.91\% in prefill and decode, respectively. Compared with the no-fast-mapping variant, \toolname{} reduces latency by 51.2\% and 50.7\% in prefill and decode, respectively, because legal copies are selected using live queue, path, bank, and streaming pressure. Compared with the no-HBM-tier variant, \toolname{} reduces decode latency by 18.4\% and repeated streaming bytes by 44.0\%. These checks support the same conclusion as Figure~\ref{fig:eval-ablation}: \toolname{}'s benefit comes from pressure-aware residency and token assignment, not from extra copy capacity alone.

\subsection{Sensitivity Study}
\label{sec:eval-sensitivity}

\begin{figure}[pos=t,align=\centering]
\centering
\ExplSyntaxOn\dim_set:Nn \l_fig_width_dim { \linewidth }\ExplSyntaxOff
\includegraphics[width=\columnwidth,trim=0 4 0 4,clip]{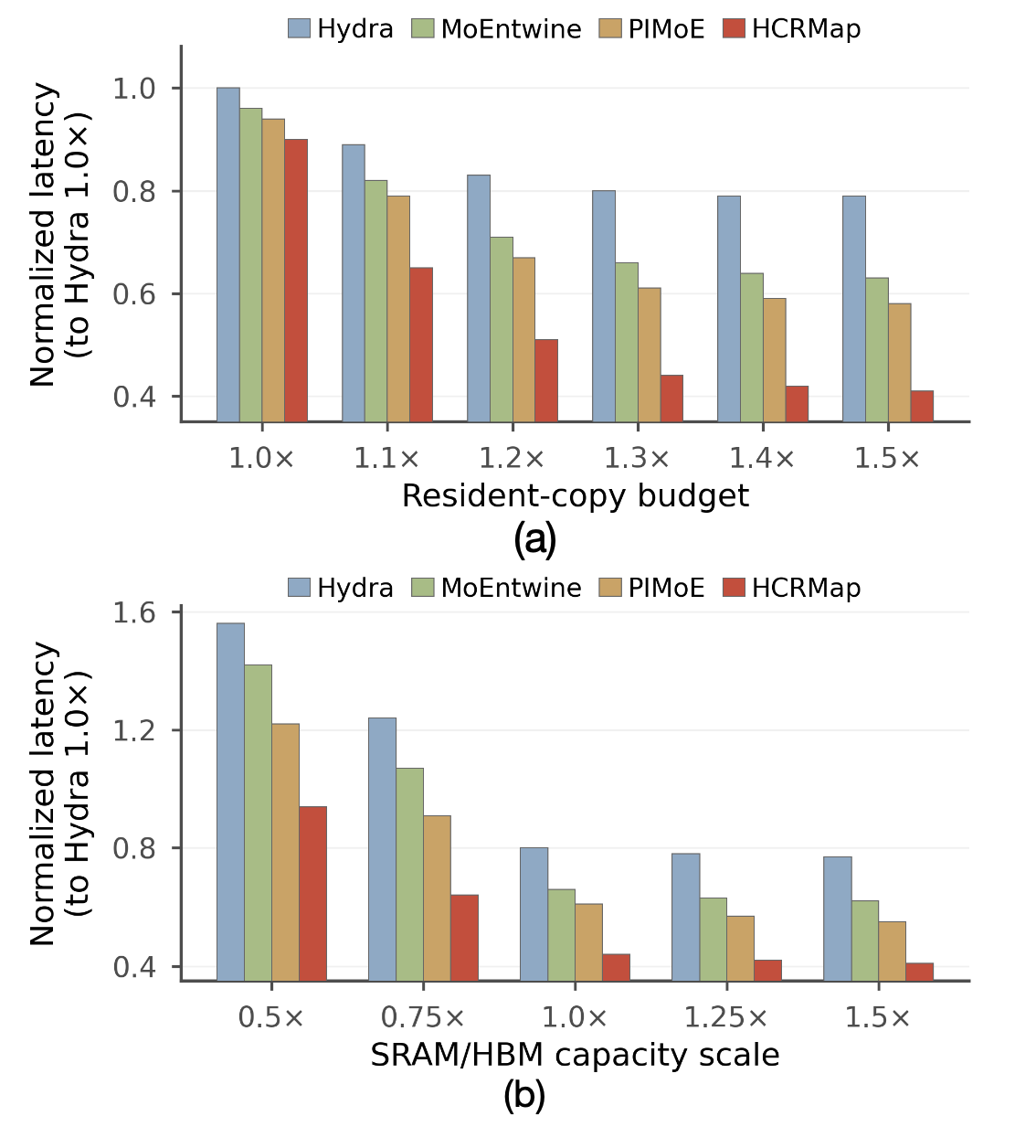}
\vspace{-4pt}
\caption{Sensitivity study: (a) resident-copy budget; (b) SRAM/HBM-capacity scale. Lower is better.}
\label{fig:eval-residency-budget}
\end{figure}

Figure~\ref{fig:eval-residency-budget} varies both resident-copy budget and SRAM/HBM capacity. In the budget sweep, moving from one copy to a modest extra-copy budget sharply reduces latency before the benefit saturates. This supports the claim that a small number of useful hot-expert execution slots removes the single-copy queueing bottleneck. In the capacity sweep, reducing SRAM/HBM capacity makes fixed replication increasingly fragile because resident copies become harder to retain without increasing memory pressure. The pressure-aware \toolname{} controller avoids harmful promotions when near-tier capacity is scarce.

\begin{figure}[pos=t,align=\centering]
\centering
\includegraphics[width=\columnwidth,trim=0 0 0 0,clip]{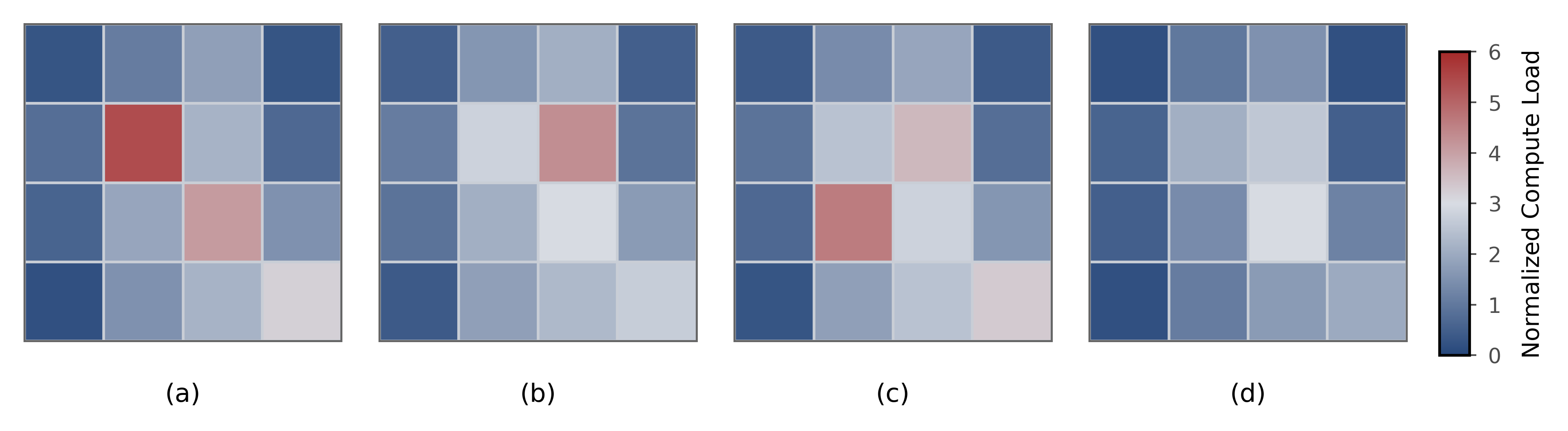}
\vspace{-4pt}
\caption{Spatial compute-pressure heatmaps: (a) Hydra; (b) MoEntwine; (c) PIMoE; (d) \toolname{}.}
\label{fig:eval-memory-sensitivity}
\end{figure}

Figure~\ref{fig:eval-memory-sensitivity} compares spatial compute-pressure heatmaps under Hydra, MoEntwine, PIMoE, and \toolname{}. All panels use the same \(4\times4\) chiplet-grid layout and the same color scale, with the colorbar reporting normalized compute load. The first three policies form localized hot spots in different grid regions, whereas \toolname{} produces a smoother and less concentrated pressure distribution.

\FloatBarrier

\subsection{Metadata Footprint}
\label{sec:eval-overhead}

After the performance, ablation, and sensitivity studies, we quantify the metadata footprint needed to support the evaluated runs.

\textbf{Metadata footprint.} Directory entries, version state, pressure counters, and pending migration buffers are modeled as accelerator metadata and are included in the capacity reservation term $M_{t,c,b}$ in Equation~(1). Directory entries are bounded at 64 B per replica, directory storage stays below 2 MB for the largest evaluated directory, pressure counters require less than 4 KB, and at most one replica is moved per decision window with the pending migration region bounded by one model-specific expert object and $U^{\max}=1$.

\FloatBarrier

\section{Conclusion}
\label{sec:conclusion}

This paper identifies hot-expert residency as a central runtime control problem in \threed{} MoE serving. Dynamic routing creates hot experts whose weights, token traffic, compute queues, and memory-bank pressure interact across the package. \toolname{} introduces a multi-level hot-expert residency substrate and a two-timescale pressure-aware mapper that jointly controls cross-tier promotion/demotion and token assignment in the routed MoE-FFN path. LEGOSim evaluation shows that reducing routed-expert queueing, memory service, and migration pressure lowers end-to-end MoE inference latency. In the matched-capacity comparison, pressure-aware placement and token assignment reduce normalized end-to-end latency by 24.95\% and 23.91\% in prefill and decode, respectively, and improve EDP by 19.94\% and 19.86\%. In the unified LEGOSim comparison against reproduced Hydra, MoEntwine, and PIMoE baselines on the same \threed{} substrate, \toolname{} reduces end-to-end latency by 34.5\% and 33.1\% in prefill and decode, respectively, showing the combined value of bounded multi-level residency and pressure-aware token assignment.

\section*{CRediT authorship contribution statement}
Yongqin Zhang: Writing -- original draft, Visualization, Software, Methodology, Investigation, Conceptualization, Writing -- review \& editing, Supervision.

\section*{Declaration of competing interest}
The author declares that there are no known competing financial interests or personal relationships that could have appeared to influence the work reported in this paper.



\section*{Data availability}
No data was used for the research described in the article.


\bibliographystyle{elsarticle-num}
\bibliography{references}

\end{document}